\documentclass[Afour,sageh,times]{sagej}
\usepackage{moreverb,url}
\usepackage[colorlinks,bookmarksopen,bookmarksnumbered,citecolor=red,urlcolor=red]{hyperref}
\newcommand\BibTeX{{\rmfamily B\kern-.05em \textsc{i\kern-.025em b}\kern-.08em
T\kern-.1667em\lower.7ex\hbox{E}\kern-.125emX}}
\usepackage{mathbbol}
\usepackage{amssymb}             
\usepackage{physics}
\usepackage{mathtools}
\usepackage{multirow}

\usepackage[ruled,vlined,noend]{algorithm2e}
\usepackage[noend]{algpseudocode}

\DeclareSymbolFontAlphabet{\amsmathbb}{AMSb}%

\DeclareMathOperator*{\argmin}{arg\,min}

\def\volumeyear{2023}

\usepackage[usernames,dvipsnames]{xcolor}

\begin{document}

\title{Sim2Real Neural Controllers for Physics-based Robotic Deployment of Deformable Linear Objects}

\author{Dezhong Tong\affilnum{1}, Andrew Choi\affilnum{2}, Longhui Qin\affilnum{1, 5}, Weicheng Huang\affilnum{1, 5}, Jungseock Joo\affilnum{3, 4}, and M. Khalid Jawed\affilnum{1}}

\affiliation{\affilnum{1} Department of Mechanical \& Aerospace Engineering, University of California, Los Angeles, CA 90024, USA\\
\affilnum{2} Department of Computer Science, University of California, Los Angeles, CA 90024, USA\\
\affilnum{3} Department of Communication, University of California, Los Angeles, CA 90024, USA \\
\affilnum{4} NVIDIA Corporation, Santa Clara, CA 95051, USA \\
\affilnum{5} School of Mechanical Engineering, Southeast University, Nanjing 211189, China}

\corrauth{Longhui Qin, M. Khalid Jawed}

\email{lhqin@seu.edu.cn, khalidjm@seas.ucla.edu}

\begin{abstract}
Deformable linear objects (DLOs), such as rods, cables, and ropes, play important roles in daily life. However, manipulation of DLOs is challenging as large geometrically nonlinear deformations may occur during the manipulation process. This problem is made even more difficult as the different deformation modes (e.g., stretching, bending, and twisting) may result in elastic instabilities during manipulation. In this paper, we formulate a physics-guided data-driven method to solve a challenging manipulation task -- accurately deploying a DLO (an elastic rod) onto a rigid substrate along various prescribed patterns. Our framework combines machine learning, scaling analysis, and physical simulations to develop a physics-based neural controller for deployment. We explore the complex interplay between the gravitational and elastic energies of the manipulated DLO and obtain a control method for DLO deployment that is robust against friction and material properties. Out of the numerous geometrical and material properties of the rod and substrate, we show that only three non-dimensional parameters are needed to describe the deployment process with physical analysis. Therefore, the essence of the controlling law for the manipulation task can be constructed with a low-dimensional model, drastically increasing the computation speed. The effectiveness of our optimal control scheme is shown through a comprehensive robotic case study comparing against a heuristic control method for deploying rods for a wide variety of patterns. In addition to this, we also showcase the practicality of our control scheme by having a robot accomplish challenging high-level tasks such as mimicking human handwriting, cable placement, and tying knots. 
\end{abstract}


\maketitle

\begin{funding}
This research was funded in part by the National Science Foundation under award numbers OAC-2209782, CMMI-2101751, CAREER-2047663, and IIS-1925360.
\end{funding}

\begin{dci}
The author(s) declared no potential conflicts of interest with respect to the research, authorship, and/or publication of this article.
\end{dci}

\section{Introduction}
Intelligent manipulation of deformable objects, such as ropes and cloth, is necessary for beneficial and ubiquitous robots. 
As most objects in the practical world are non-rigid, endowing robots with proper manipulation skills for deformable objects has enormous humanitarian and economic potential. 
Some examples include robotic surgical suturing~\citep{sen2016automating, stefanidis2010robotic}, wire management~\citep{she2021cable}, laundry folding~\citep{miller2012geometric}, and caregiving for elderly and disabled communities~\citep{kapusta2019personalized, clegg2018learning, yu2017haptic, erickson2018deep, pignat2017learning}. 
However, given the large and geometrically nonlinear deformations of deformable objects, it is difficult to obtain an obvious mapping from the observations of those manipulated objects to a concrete robotic manipulation scheme.
Therefore, developing accurate and effective strategies for manipulating deformable objects is still an open research problem.

Among various deformable objects, deformable linear objects (DLOs), which include elastic rods and the extensions of rods-like structures, e.g., cables, ropes, rods, and wires~\citep{sanchez2018robotic}, have attracted significant research interest due to their widespread industrial and domestic applications. In this article, we focus on the category of rod-like structures and adopt the term DLO to refer to those solid elongated objects.
DLOs usually possess extremely complicated nonlinearity due to the coupling of their multiple deformation modes: stretching, bending, and twisting.
Given the practicality and difficulty of manipulating DLOs, there is a growing need for robust and effective methods to manipulate DLOs.

\begin{figure*}[t!]
\centerline{\includegraphics[width =\textwidth]{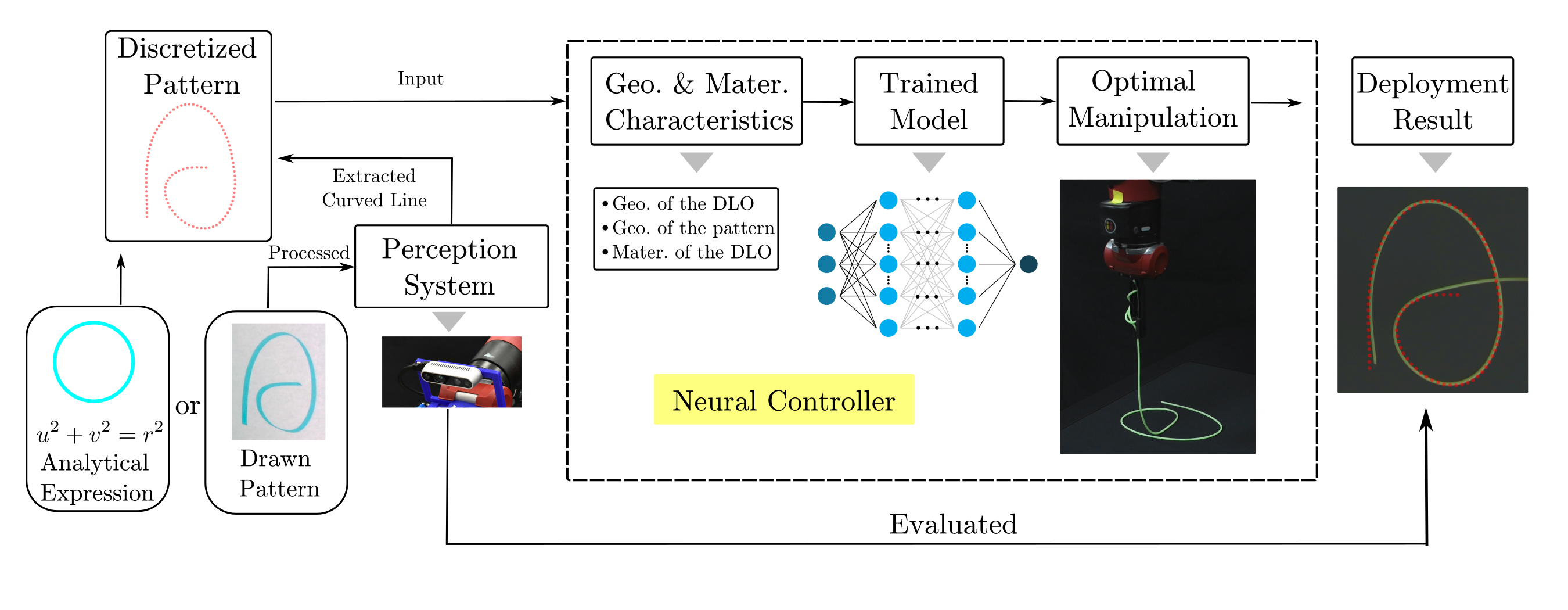}}
\caption{A full end-to-end pipeline for deploying a DLO with a sim2real physics-based deployment scheme. The pipeline begins by discretizing the DLO pattern, which can be obtained through user input via an analytical expression or a hand-drawn pattern scanned by a perception system~\citep{choi2023mBest}. A neural controller trained entirely from simulation then generates an optimal manipulation path for deploying the pattern, taking into account the shape of the pattern as well as the geometrical and material properties of the DLO. Finally, the deployment result is evaluated using an Intel RealSense camera positioned to provide a top-down view of the pattern to assess the accuracy of the deployment.}
\label{fig::first}
\end{figure*}

Prior works on manipulating DLOs can be divided into two categories. 
The first involves robots attempting to manipulate DLOs to satisfy some high-level conditions without controlling the exact shapes of DLOs. This includes knot tangling/untangling~\citep{wakamatsu2006knotting,saha2007manipulation}, obstacle avoidance~\citep{mcconachie2020manipulating, mitrano2021learning}, following guidance and insertion~\citep{she2021cable, zhu2019robotic}, etc. 
The second category involves robots attempting to precisely control the exact shape of the DLOs.
For this task, a key challenge is formulating a mapping between the robot's motions and the shape of the manipulated DLO~\citep{nair2017combining,takizawa2015method,lv2022dynamic}.
In this article, we look into how to design a manipulation scheme for controlling the shape of elastic rods through deployment, which involves manipulating one end of DLO in a way that gradually lays the DLO on a substrate in a desired pattern with superhuman accuracy, sufficient efficiency, and strong robustness. The full end-to-end pipeline of our physics-based deployment scheme is shown in Fig.~\ref{fig::first}.
In addition to achieving precise shape control, we show our control method can be used to solve high-level tasks such as reproducing human writing with a deployed DLO, cable placement, and knot tying.

\begin{figure*}[t!]
\centerline{\includegraphics[width =\textwidth]{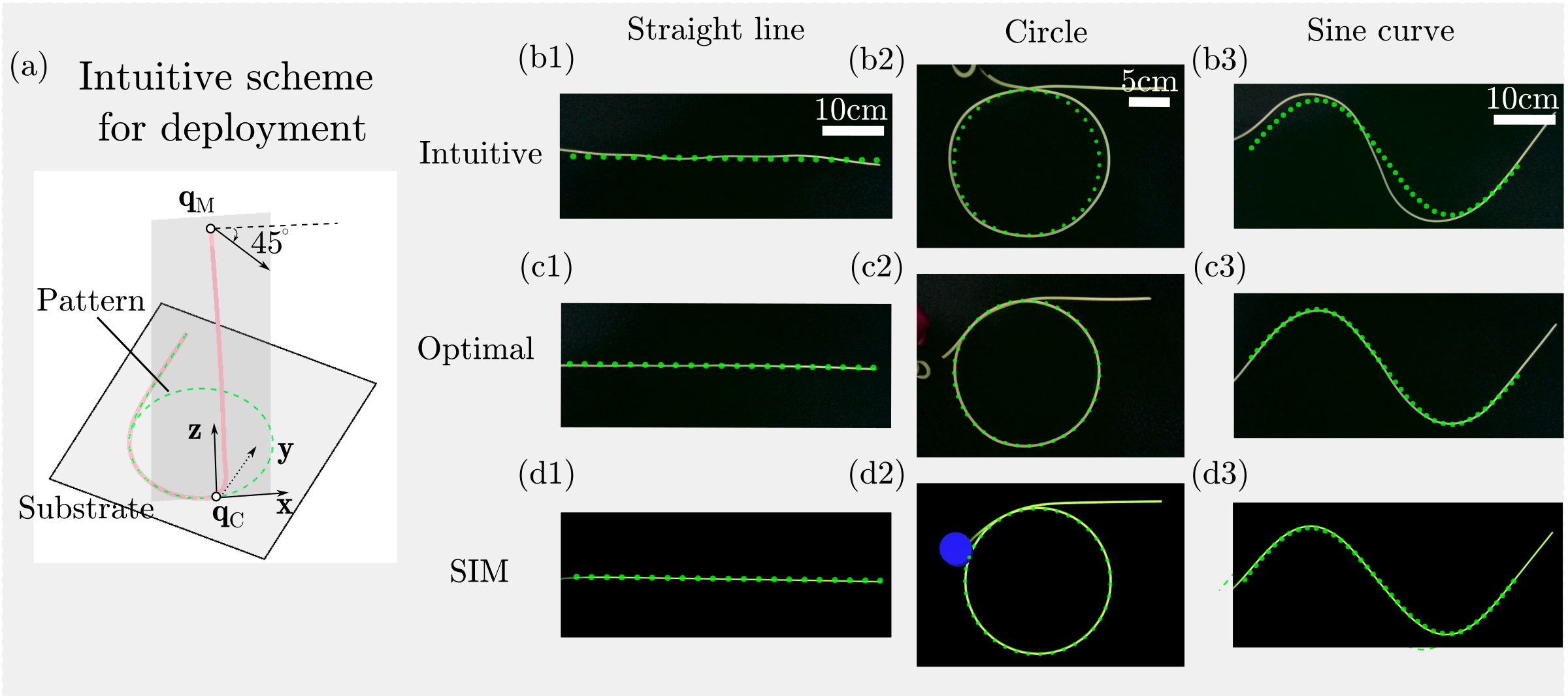}}
\caption{
(a) Schematic of the intuitive control method from \citet{takizawa2015method}. A DLO is being deployed along a circular pattern shown in dashed green. During the deployment process, the manipulated position $\mathbf q_M$ deploys along the tangent of the pattern $\mathbf x$ in a downward 45-degree angle with respect to the $y$-axis. The $x$-$z$-plane is shown in opaque gray.
In addition, a comparison of experimental results between the (b) intuitive control method, (c) our designed optimal control method, (d) and simulation results using the optimal control method for the patterns of straight line, circle, and sine curve are shown.
Note the effects of forgoing the influence of nonlinear geometric deformations in the intuitive deployment scheme's failure to follow simple patterns.}
\label{fig::introduction}
\end{figure*}

\subsection{Deployment of DLOs}
Deploying DLOs is instrumental in the practical world, e.g., drawing or writing on cakes with icing~\citep{sun2015review}, deploying marine cables~\citep{whitcomb2000underwater}, depositing carbon nanotubes~\citep{geblinger2008self}, and melting electrospinning for advanced manufacturing~\citep{teo2006review}.
Therefore, a concrete and applicable deployment scheme is a perfect solution to the shape control problem of DLOs. 

Now a natural question arises: how to deploy a DLO along a prescribed pattern accurately on a substrate?
Intuitively, we can assume that during the deployment process, the manipulated end $\mathbf q_M$ is directly above the contact point $\mathbf q_C$ and that the gripper's decreasing distance along the negative $z$-axis is equal to the added deployed length on the substrate.
However, this deployment strategy does not take into consideration the nonlinear geometric deformations of the manipulated DLO and therefore, results in a poor quality deployment as illustrated by later experimental results.
A schematic of the intuitive deployment method inspired by \citet{takizawa2015method} can be observed in Fig.~\ref{fig::introduction}(a).

In this paper, we propose a framework that combines physically accurate simulation, scaling analysis, and machine learning to generate an optimized control scheme capable of deploying solid rod-like structures, which we refer to as DLOs, along any feasible pattern. Our control scheme does not currently incorporate energy dissipation from manipulations with DLOs such as viscous threads, as our physical-based simulation is based on the rod model. However, the controlling scheme can be adapted by adjusting the physical-based simulation in our combined framework to include these factors.
We validate the scheme with various DLOs ( e.g., elastic rods, rope, and cable) in robotic experiments. 
The usage of physically accurate numerical simulations not only allows us to incorporate physics into our manipulation scheme but also results in full sim2real realization.
Scaling analysis allows us to formulate the problem with generality using non-dimensional parameters, resulting in a control scheme robust against the material properties of the manipulated rods.
Finally, machine learning allows us to train a neural network to model the controlling rules of deployment in a data-driven fashion. 
The high inference speed of our neural controller makes real-time operation feasible.

Our main contributions are as follows: 
(1) we formulate a solution to the DLO shape control problem through deployment with a physically robust scheme that leverages scaling analysis, resulting in generality against material, geometric, and environmental factors (friction);
(2) we train a neural network (NN) with non-dimensional simulation data to serve as a fast and accurate neural controller for optimal manipulations of deployment tasks. The trained mechanics-based NN solver has remarkable efficiency and sufficient accuracy when compared to a numerical solver; and 
(3) we demonstrate full sim2real realization through an extensive robotic case study demonstrating our control method's success for various practical deployment patterns with various DLOs on different substrates. In addition, we showcase the utility of our control scheme for complex high-level applications such as mimicking human handwriting, managing cables, and tying different knots.

Moreover, we have released our source codes and supplementary videos\footnote{\label{github}See \url{https://github.com/StructuresComp/rod-deployment}.}.

\subsection{Overview}
The remainder of the article is organized as follows: we begin with a literature review related to robotic DLO shape control in Sec.~\ref{sec:related-work}.
The formulation of the physics-based numerical model is discussed in Sec.~\ref{sec:numerical-framework-and-physical-analysis}, where we also formulate the deployment problem with scaling analysis.
In Sec.~\ref{sec:optimization-and-deep-learning}, we analyze the nonlinearity of the deployment in detail and show how to discover optimal robot manipulation through numerical simulation. 
In addition, a learning framework is formulated to obtain a fast, generalized motion planning solution. 
Next, in Sec.~\ref{sec:pereception-system-and-robotic-system}, we introduce our overall robotic system, including perception and motion planning modules. 
Experimental results and analysis for different deployment cases, including writing letters and tying a knot, are given in Sec.~\ref{sec:experiments}.
Finally, we provide concluding remarks and discuss future research avenues in Sec.~\ref{sec:conclusion}.

\section{Related Work}
\label{sec:related-work}
Constructing a mapping relationship from observations of a manipulated DLO to the robot's action space is the primary basis of controlling DLOs. 
To uncover this mapping relationship, prior works usually implemented models to predict or perception systems to observe the deformations of DLOs under various manipulations. 
Manipulation schemes are then generated based on the predicted or sensed data. 
Therefore, model-based and perception-based methods can be considered two of the main categories for tackling manipulation problems of deformable objects. 
Due to the outstanding performance of machine learning algorithms for processing and generalizing data from models and perceptions, learning-based approaches have become another mainstream solution. 
In fact, many prior works take advantage of a combination of these three methods to develop hybrid schemes for different manipulation tasks.
Here, we carry out a systematic review of prior scholarly contributions that have utilized techniques based on the three delineated categories to manipulate DLOs and other deformable objects.

Perception-based approaches involve utilizing sensors such as tactile sensors~\citep{she2021cable} and cameras~\citep{tang2018framework, yan2020self, lee2014unifying, maitin2010cloth} to generate motions based on detected deformations. 
While sensors can capture the deformations as the manipulation proceeds, perception-based methods are usually not robust against the material and geometrical differences of the manipulated objects.
In \citet{tang2018framework}, a learning-based perception framework is presented based on the Coherent Point Drift algorithm, which is able to register states of manipulated DLOs with captured images. 
\citet{yan2020self} developed state estimation algorithms for DLOs based on images so that a robot can perform pick-and-place manipulation on the detected configuration. 
However, those perception systems based on cameras fail to extract accurate results when occlusions happen. 
To overcome this shortcoming, tactile sensors have become prevalent in the robotics community. 
For example, \citet{she2021cable} implements GelSight, a force feedback tactile sensor, to perform robotic cable management.
Since sensing data by itself cannot predict future deformations of the manipulated objects, pure perception-based methods are typically insufficient for complex deformable material manipulation tasks.

Model-based methods usually construct a physically accurate model to predict the behavior of manipulated DLOs. 
Multiple methods exist for modeling DLOs~\citep{yin2021modeling, sanchez2018robotic}. 
A simple and widely-used model, mass-spring systems, are often used to model deformable objects including ropes~\citep{schulman2013case, kita2011clothes, macklin2014unified}, fabrics~\citep{macklin2016xpbd, guler2015estimating}, etc. 
However, due to the simplification of mass-spring systems, such models usually suffer from inaccuracies when undergoing large deformations and lack of physical interpretability.
Position-based dynamics is another type of modeling method that usually represents DLOs as chains of rigid bodies~\citep{servin2008rigid, terzopoulos1994dynamic, muller2007position} and introduces constraints between the positions of those rigid bodies to simulate deformations. 
Though this method is straightforward and fast, physical interpretability is also lacking. 

Finite element methods (FEM) are also popular for modeling deformable objects~\citep{haouchine2018vision, kaufmann2009flexible, buckham2004development}. 
However, FEM usually requires considerable computation resources and is hardly suitable for online predictions. 
More recently, fast simulation tools from the computer graphics community have attracted researchers' attention. 
For example, Discrete Elastic Rods (DER)~\citep{bergou2008discrete, bergou2010discrete} has arisen as a robust and efficient algorithm for simulating flexible rods. 
\citet{lv2022dynamic} used DER as a predictive modeling tool and achieved promising performance in DLO manipulation tasks. 
Though various ways to model deformable objects exists, each has their respective strengths and weaknesses and often possesses a trade-off between computational efficiency and accuracy.

Finally, learning-based approaches have become prevalent as they are capable of not only predicting the shape of the deformable object but also higher-level information such as forces \citep{choi2023deep}. 
Most prior works use human demonstrations or robot explorations to train controlling policies for different tasks. 
\citet{nair2017combining}, \citet{sundaresan2020learning}, and \citet{lee2021sample} fed human-made demonstrations to robots for learning control policies for shape control and knot-tying. 
Due to the tedium of constructing manual demonstrations, some researchers take advantage of the robots' automation to learn a policy purely from robotic exploration~\citep{yu2022global, wang2019learning}. 
To acquire training data more efficiently, researchers have also looked into training policies purely from simulation~\citep{matas2018sim}. 
Although learning-based methods have shown promising performance for manipulating deformable objects, the trained policies are often only valid for specific tasks whose state distribution matches that of the training set. 
In other words, learning-based approaches often fail when parameters such as the material and geometrical properties of the manipulated object change.

More relevant to the deployment task itself, ~\citet{takizawa2015method} implemented the intuitive control method shown in Fig.~\ref{fig::introduction}(a) for controlling the shape of a rope to make a clove hitch knot. 
They achieve a success rate of 60\% but require empirical hardcoded adjustments to their controlling scheme, indicating the intuitive approach's unsuitability for extreme precision deployment.
Additionally, ~\citet{lv2022dynamic} uses a precise physical numerical model to predict the DLO's configuration during deployment.
However, they use a trial-and-error method to exhaustively solve the optimal deployment path, which is computationally expensive and slow. 

Although the three discussed types of methods are suitable to be combined when solving deformable manipulation problems given the complementariness of their pros and cons, how to develop a combined approach to take advantage of different types of approaches is still an open problem in the robotic community. 
We find that combining physically accurate simulations and machine learning can endow the learned model with excellent accuracy from the simulations and real-time performance because of the inference speed of the neural network. 
In addition, scaled physics analysis, which is a vital tool from the mathematical physics community, is valuable for augmenting the model with high generality.

In this article, we show how physical analysis can extract the true contributing factors of the deployment problem and how a learning-based approach can generalize the information from physics to offer real-time computation speed for the manipulation task.

\section{Numerical Framework and Physical Analysis}
\label{sec:numerical-framework-and-physical-analysis}
\begin{figure}[!t]
\centerline{\includegraphics[width =\columnwidth]{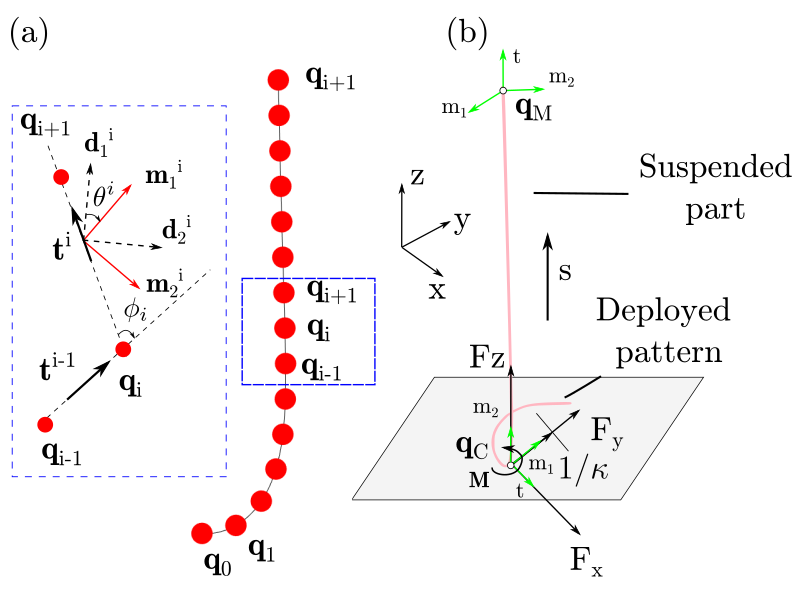}}
\caption{(a) Discrete diagram of the centerline of a DLO and relevant notations; (b) schematic of deploying a DLO along a prescribed pattern.}
\label{fig::schematic}
\end{figure}

In this section, we first discuss the numerical framework for studying the nonlinear behaviors of the DLO during deployment. 
Then, we extract the main controlling parameters for the analyzed system with Buckingham\rq{}s $\pi$ theorem. 

\subsection{Discrete Differential Geometry (DDG)-based Numerical Framework}
\label{sec::DDG}
To simulate DLOs, we use a DDG-based simulator -- Discrete Elastic Rods (DER)~\citep{bergou2008discrete, bergou2010discrete} -- whose physical accuracy has been validated in many various scenarios such as flagella motions~\citep{jawed2015propulsion}, knot tying~\citep{choi_imc_2021, tong2023snap, tong2023fully}, rod coiling~\citep{jawed2014coiling}, and elastic buckling in structures~\citep{tong2021automated}.

As shown in Fig.~\ref{fig::schematic}(a), the centerline of a DLO can be discretized into $N+1$ nodes $[\mathbf q_0, \mathbf q_1, ..., \mathbf q_N]$ ($\mathbf q_i \in \mathbb R^3$) and $N$ edges $[\mathbf e^0, \mathbf e^1, ..., \mathbf e^{N-1}]$ ($\mathbf e^i = \mathbf q_i - \mathbf q_{i-1}$). 
In this section, node-relevant quantities are denoted with subscripts, e.g., $\mathbf q_i$, while edge-relevant quantities are denoted with superscripts, $\mathbf e^i$. 
Each edge $\mathbf e^i$ possesses two orthogonal frames: a reference frame $[\mathbf d_1^i, \mathbf d_2^i, \mathbf t^i]$ and a material frame $[\mathbf m_1^i, \mathbf m_2^i, \mathbf t^i]$. 
The material frame, which captures the rotation of the centerline of the DLO, can be obtained by rotating the reference frame by a rotation angle $\theta^i$ with respect to the shared director $\mathbf t^i$. 
The reference frame is arbitrarily initialized at the initial time $t = 0$s and is updated between time steps using time parallel transport~\citep{bergou2010discrete}. 
The following DOF vector of size $(4N+3)$ is constructed to capture all the deformations of the rod:
\begin{equation}
    \mathbf q = \left[\mathbf q_0, \theta^0, \mathbf q_1, ..., \mathbf q_{N-1}, \theta^{N-1}, \mathbf q_N \right]^T,
\end{equation}
where $T$ is the transpose operator.

Based on DER~\citep{bergou2008discrete, bergou2010discrete}, the deformations of a DLO can be divided into three modes, each corresponding to a distinct type of elastic energy: stretching, bending, and twisting.
Using the formulations of these elastic energies in DER, we can outline the equations of motion (EOM) we must solve at each time step.

First, we write down the formulation of stretching energy:
\begin{equation}
    E_s = \frac{k_s}{2} \sum_{i=0}^{N-1}  \left( 1 - \frac{ \lVert \mathbf{e}^i \rVert}{\lVert \tilde{\mathbf{e}}^i \rVert} \right)^2 \lVert \tilde{\mathbf{e}}^i \rVert,
    \label{eq:stretching}
\end{equation}
where $k_s$ is the stretching stiffness and $\lVert \tilde{\mathbf e}^i \rVert$ is the undeformed length of the $i-$th edge. 
Note that we assume that the manipulated rod is of an isotropic linear elastic material in this manuscript. 
Hereafter, all quantities with $\tilde{()}$ refer to their resting undeformed value.

Next, the bending energy is outlined as
 \begin{equation}
    E_b = k_b \sum_{i=1}^{N-1} \frac{
    (\boldsymbol \kappa_i - \tilde{\boldsymbol \kappa}_i)^T (\boldsymbol \kappa_i - \tilde{\boldsymbol \kappa}_i)
    }{\lVert \tilde{\mathbf e}^i \rVert + \lVert \tilde{\mathbf e}^{i-1} \rVert} ,
    \label{eq:bending}
\end{equation}
where $k_b$ is the bending stiffness, and $\boldsymbol{\kappa}, \hat{\boldsymbol{\kappa}} \in \mathbb R^3$ are the deformed and undeformed curvature vectors, respectively. 
Here, the relationship between the turning angle $\phi_i$ and curvature $\boldsymbol \kappa_i$ is given as $2\tan(\phi_i/2) = \lVert \boldsymbol{\kappa}_i\rVert$. The illustration of turning angle $\phi_i$ can be seen in Fig.~\ref{fig::schematic}(a).
Note that we also assume that the resting undeformed shape of the rod is straight, i.e., $\tilde{\phi}_i = 0$ in our study.

Finally, the twisting energy is
\begin{equation}
    E_t = k_t \sum_{i=1}^{N-1} \frac{(\tau_i - \tilde{\tau}_i)^2}{\Vert \bar {\mathbf e}^i \Vert + \Vert \bar {\mathbf e}^{i-1} \Vert},
    \label{eq:twisting}
\end{equation}
where $k_t$ is the twisting stiffness, $\tau_i = \theta^i - \theta^{i-1} + \Delta \tau_i^{\textrm{ref}}$ is the discrete twist, $\tilde {\tau}_i$ is the natural twist, and $\Delta \tau_i^{\textrm{ref}}$ is the angular difference between the reference frames on edges $\mathbf e^{i-1}$ and $\mathbf e^i$. For our DLOs, we presume $\tilde {\tau}_i$ to be zero.

With Eqs.~\ref{eq:stretching}, \ref{eq:bending}, and~\ref{eq:twisting}, the internal forces of the rod can be obtained as
\begin{equation}
    \mathbf F^{\textrm{int}} = - \frac{\partial(E_s + E_b + E_t)}{\partial \mathbf q}.
    \label{eq:internal}
\end{equation}
We can then construct the equations of motion implicitly based on Newton's second law:
\begin{subequations}
	\begin{align}
	\begin{split}
	& \mathcal{R}(\mathbf q) \equiv  \frac{\mathbb{M}}{\Delta t} \left( \frac{\mathbf{q} (t_{i+1}) - \mathbf{q} (t_i)}{\Delta t} - \dot{\mathbf{q}} (t_i) \right) - \mathbf{F}^{\textrm{int}} - \mathbf{F}^{\textrm{ext}} = 0,
	\end{split}\\
	\begin{split}
	& \dot{\mathbf{q}} (t_{i+1}) = \frac{\mathbf{q} (t_{i+1}) - \mathbf{q} (t_i)}{\Delta t},
	\end{split}
	\end{align}
	\label{eq:EOM}
\end{subequations}
where $\mathbb{M}$ is a square lumped mass matrix of size $4N+3$; $\mathbf F^\textrm{int}$ is a $(4N+3) \times 1$ elastic force vector (from Eq.~\ref{eq:internal}), and $\mathbf F^\textrm{ext}$ is a $(4N+3) \times 1$ external force vector. 
The $\dot{(\,)}$ operator represents the derivative of a quantity with respect to time, i.e., $\dot{\mathbf{q}} (t_i)$ is the velocity vector at time $t_i$. 
Note that the subscript in Eq.~\ref{eq:EOM} is the time stamp. 
By solving Eq.~\ref{eq:EOM} with Newton's method, the nonlinear geometric deformation of the manipulated rod over time can be simulated accurately.

\subsection{Physical Analysis and Controlling Rule Construction}
\label{sec::physicalAnalysis}

When manipulating DLOs, we should consider their geometrically nonlinear deformations. 
Moving forward, $\hat{\mathbf x}$, $\hat{\mathbf y}$, and $\hat{\mathbf z}$ refer to the unit directors of the coordinate system defined by the connective node $\mathbf q_C$ shown in Figs.~\ref{fig::introduction}(a) and \ref{fig::schematic}(b).

As shown in Fig.~\ref{fig::schematic}(b), when a DLO is being deployed along a prescribed pattern on a rigid substrate, it can be divided into two parts: a deployed part on the substrate and a suspended part that does not contact the substrate. 
Here, we presume the pattern on the substrate is fixed since the DLO should ideally be deployed along the prescribed pattern. Therefore, the unknown deformations only exist in the suspended part.

\subsubsection{Solving the Suspended Part}
To capture the deformations of the suspended part, we introduce some quantities to assist our analysis. First, we define $\mathbf{q}(s)$ to describe the position of the manipulated DLO\rq{}s centerline, where $s$ is the arc length along the DLO\rq{}s centerline. Then, a material frame $\mathbf{m}(s) = [\mathbf{m}_1, \mathbf{m}_2, \mathbf{t}] \in SO(3)$ is attached along the DLO to capture the DLO's rotation, where $\mathbf{t} = \dv{\mathbf q}{s}$ is the tangent of the DLO. With the help of $\mathbf{q}(s)$ and $\mathbf{m}(s)$, we can fully describe the deformed configuration of the suspended part.

To solve the configuration of the suspended part, we can treat the suspended part as an independent DLO starting from the connective node $\mathbf{q}_C$ to the manipulated node $\mathbf{q}_M$. Here, $\mathbf{q}_C = \mathbf{q}(0)$ is the connective node connecting the deployed part and the suspended part. Given the continuity of the manipulated DLO, the curvature vector $\boldsymbol{\kappa}$ at $\mathbf{q}_C$ can be obtained from the prescribed pattern, where the magnitude of $\boldsymbol{\kappa}$ is denoted as $\kappa$.
The manipulated end grasped by the robot is then $\mathbf{q}_M = \mathbf{q}(l_s)$, where $l_s$ is the total curve length of the suspended part. 
Deployment of the pattern is then carried out purely by controlling $\mathbf q_M$.
Since Eq.~\ref{eq:EOM} implies that the DLO\rq{}s configuration $\mathbf{q}(s)$ and $\mathbf{m}(s)$ can be solved when boundary conditions are determined, we can write down the governing equations for the suspended part as
\begin{equation}
\begin{aligned} 
\mathcal{R}(\mathbf q) & = 0,\\
 \textrm{s.t.} \quad  \mathbf q(l_s) &= \mathbf q_M, \ \ \ \mathbf R = \mathbf m(l_s) \mathbf m(0)^T,\\
 \mathbf q(0) &= \mathbf q_C, \ \ \ \dv{\mathbf q(0)}{s} = \mathbf t(0), \ \ \ \dv{\mathbf t(0)}{s} = \kappa \hat{\mathbf y}, \\
\end{aligned}
\label{eq::BVP}
\end{equation}
where $\mathbf q_M$ is the position and $\mathbf R$ is the orientation of the manipulated end with respect to the connective node $\mathbf q_C$.
Note that the position of the connective node $\mathbf q_C$, tangent $\mathbf t(0)$, and curvature vector $\kappa \hat{\mathbf{y}}$ can be determined from the deployed pattern, where $\hat{\mathbf y}$ is the unit vector illustrated in Fig.~\ref{fig::schematic}(b).
By solving Eq.~\ref{eq::BVP}, we can obtain the configuration of the suspended part for any predefined pattern and manipulated end pose.

\subsubsection{Influence of Forces and Friction}
Once the deformed configuration is known, we can now calculate the forces applied on the suspended part, which is key to hyper-accurate control of the DLO. 
We denote the external forces $\mathbf F^\textrm{ext}=F_x \hat{\mathbf x}+F_y \hat{\mathbf y} +F_z \hat{\mathbf z}$ and twisting moment $M(0)$ applied on the suspended part from the connective node $\mathbf q_C$. 
Here, the moment $M$ is a function of arc length $s$; for example, $M(s)$ is the twisting moment applied on the manipulated end.
The quantities $\mathbf F^\textrm{ext}$  and $M(0)$ are relevant with the friction coefficient $\mu$ between the substrate and the rod, and $\mu$ is an unknown and uncontrollable environment factor. In addition, the quantities $\mathbf F^\textrm{ext}$  and $M(0)$ also influence the tangent $\mathbf t(0)$ at the connective node $\mathbf q_C$ because of Newton's third law.
Therefore, we must minimize quantities $\mathbf F^\textrm{ext}$ and $M(0)$ to achieve an optimal controlling rule.

Despite the optimal controlling rule minimizing the influence of friction, it is still worth clarifying the significance of friction within this context. 
Though we make the strong assumption that the deployed pattern remains fixed during deployment, this is only upheld if the following relation is satisfied for the deployed segment:
\begin{equation}
\begin{aligned} 
k_b \kappa'' & \leq \mu_s \rho A g,\\
\end{aligned}
\label{eq::bendingFriction}
\end{equation}
where $k_b$ is the bending stiffness of the rod, $\kappa''$ is the second derivative of $\kappa$ with respect to the arc length $s$ ($\kappa'' = \dv[2]{\kappa}{s}$), $\mu_s$ is the static friction coefficient, $\rho$ is the volumetric density of the rod, $A$ is the cross-sectional area, and $g$ is the gravitational acceleration. 
Eq.~\ref{eq::bendingFriction} is derived by analyzing an arbitrary finite element of the deployed pattern with a clamped-end Euler-Bernoulli beam model.
Clearly, friction plays a crucial role in the deployment process. 

As a result, our designed optimal deployment strategy maintains a reliance on adequate friction for effective execution while the scheme mitigates external tangential forces apart from the essential friction on the substrate. Consequently, the scheme necessitates only a modest static friction coefficient between the substrate and the manipulated DLO. 

\subsubsection{Computing Optimal Grasp}
In addition to the minimization of the external forces $\mathbf F^\textrm{ext}$ and twisting moment $M(0)$ applied on the suspended part, we set up a rule for the manipulated end: the robot end-effector should induce minimal deformations on the manipulated node $\mathbf q_M$ so that the curvature (bending deformations) at the manipulated end should be 0.
This results in the following optimization problem to compute the optimal grasp: 
\begin{equation}
\begin{aligned}
(\mathbf q_M^*, \mathbf R^*) &= \argmin \left(\lVert \mathbf F^{\textrm{ext}} \rVert^2 + \left(\frac{\lVert \mathbf M(0) \rVert}{h}\right)^2 \right) \\
 \textrm{s.t.} \quad  \dv{\mathbf q(0)}{s} &= \mathbf t(0), \ \ \ \dv{\mathbf t(0)}{s} = \kappa \hat{\mathbf y}, \\ 
 \dv[2]{\mathbf q(l_s)}{s} &= 0, \ \ \ \ \ \ \ \ \ \ \mathcal{R}(\mathbf q) = 0.
 \end{aligned}
\label{eq::optimize}
\end{equation}

By solving Eq.~\ref{eq::optimize}, optimal grasp ($\mathbf q_M^*$ and $\mathbf R^*$) can be obtained. Physical analysis tells us that a direct mapping relationship exists between the contributing factors and the optimal grasp.
Recall from Eqs.~\ref{eq:stretching},~\ref{eq:bending}, and~\ref{eq:twisting}, that stretching stiffness $k_s$, bending stiffness $k_b$, twisting stiffness $k_t$, density $\rho$, and rod radius $h$ are the primary material and geometric properties of a rod. By adding in additional geometric properties such as suspended length $l_s$ and curvature $\kappa$, the mapping relationship
\begin{equation}
\begin{aligned} 
(\mathbf q_M^*, \mathbf R^*) &= f(l_s, \kappa, k_s, k_b, k_t, h, \rho), \\
\end{aligned}
\label{eq::mapping}
\end{equation}
can be constructed where $f(\cdot)$ is a highly nonlinear (and unknown) function that describes the controlling rule. 

Note however the high input dimensionality of Eq.~\ref{eq::mapping}.
In other words, to accurately learn the mapping $f(\cdot)$, we would have to exhaustively perform large parameter sweeps for various ranges of material and geometric parameters within simulations.
This process of collecting data quickly gets out of hand due to the curse of dimensionality.
To circumvent this, we can perform scaling analysis to obtain an equivalent reduced-order mapping.

\subsubsection{Scaling Analysis via Buckingham\rq{}s $\pi$ Theorem}
In this article, we use Buckingham\rq{}s $\pi$ theorem to reduce the dimensions of the mapping $f(\cdot)$. Buckingham\rq{}s $\pi$ theorem is a fundamental principle in dimensional analysis, stating that a physically meaningful equation involving $n$ physical parameters can be expressed using a reduced set of $p = n - k$ dimensionless parameters derived from the original parameters. Here, $k$ represents the number of physical dimensions. Using Buckingham\rq{}s $\pi$ theorem allows us to obtain a reduced-order non-dimensionalized mapping $\mathcal{F}(\cdot)$ from the original function $f$:
\begin{equation}
\begin{aligned} 
(\bar{\mathbf q}_M^*, \mathbf R^*) &= \mathcal{F}(\bar l_s, \bar \kappa, \bar k_s), \\
L_{gb} &= \left(\frac{k_b}{2 \pi h^2 \rho g}\right)^{1/3}, \\
\bar k_s &= \frac{k_s L_{gb}^2}{k_b}, \\
\bar{\mathbf q}_M^* &= \frac{\mathbf q_M^* - \mathbf q_C}{L_{gb}}, \\
\bar l_s &= \frac{l_s}{L_{gb}}, \\
\bar \kappa &= \kappa L_{gb}.\\
\end{aligned}
\label{eq::nondimensionl_mapping}
\end{equation}

Hereafter, all quantities with $\bar{()} $ indicate normalized quantities. In Eq.~\ref{eq::nondimensionl_mapping}, all quantities are unitless so that the mapping relationship $\mathcal{F}(\cdot)$ maps from the unitless groups encapsulated the geometric and material properties to the unitless optimal robotic grasp. The benefit of doing such is that we reduce the dimensions of the mapping function $\mathcal{F}(\cdot )$ in Eq.~\ref{eq::mapping} and eliminate the dependence of $\mathcal{F}(\cdot)$ on the units. Note that in Eq.~\ref{eq::nondimensionl_mapping}, we do not consider the influence of the twisting stiffness $k_t$ in this article since twisting energies are minimal compared to bending and stretching. 
However, the influence of $k_t$ can also be analyzed with our proposed analysis. In the following article, we will show how to establish the nonlinear mapping function in Eq.~\ref{eq::nondimensionl_mapping}.

\section{Optimization and Deep Learning}
\label{sec:optimization-and-deep-learning}
In this section, we further analyze the optimization of the system to obtain the nonlinear mapping function in Eq.~\ref{eq::nondimensionl_mapping}. 
Given the high nonlinearity of the system, we first solve Eq.~\ref{eq::nondimensionl_mapping} with a numeric optimization solver in a data-driven way.
While doing so, we analyze the elastic instability of the system to choose the optimal robotic grasp for the deployment task. 
Afterward, we reconstruct Eq.~\ref{eq::nondimensionl_mapping} using a neural network to take advantage of its high inference speed.
This neural controller is then used by our robotic system as the controlling law to complete various deployment tasks in Sec.~\ref{sec:experiments}.

\subsection{Elastic Instability in Deployment along a Straight Line}

\begin{figure}[t!]
\centerline{\includegraphics[width =\columnwidth]{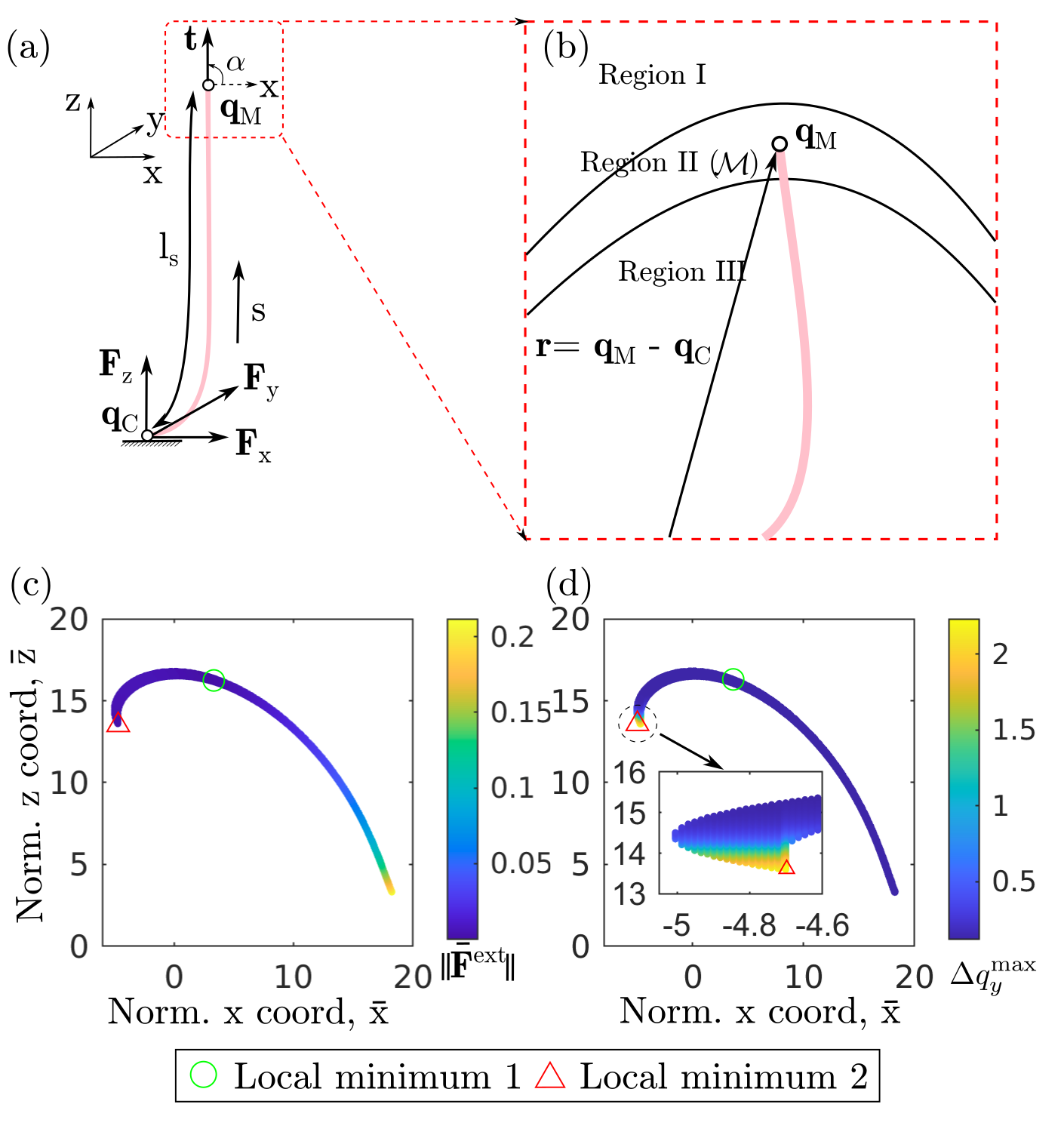}}
\caption{Schematic of a DLO manipulated in a 2D workspace (a) and its corresponding available region denoted by $\mathcal{M}$ (b). Visualization of a specific case with $\bar{l}_s = 17.68$. The force distribution is shown in (c), and (d) displays the maximum geometric deformation of the suspended part under a disturbance of $\Delta \bar{y} = 0.12$ along the $y$-axis.}
\label{fig::2Dschematic}
\end{figure}

In this section, we first take a look at an intriguing physical phenomenon: elastic instability.
Elastic instability occurs when changes in the boundary conditions cause a deformed structure to become unstable. 
When observed visually, a small geometric perturbation of the system will lead to a substantial change in configuration~\citep{timoshenko2009theory}.
An example of this can be observed when a robot employs the intuitive control method to deploy a DLO along a straight line as the rod unexpectedly adopts a curved shape on the substrate.
This observation defies our intuition as the intuitive method only manipulates the DLO in the 2D plane ($x$-$z$ plane) as illustrated in Fig.~\ref{fig::2Dschematic}(a). 
Consequently, the suspended part should ideally experience only 2D deformations within that plane, thereby avoiding significant deformations along the $y$-axis. 
On the contrary, this observed phenomenon results from the unaccounted elastic instability of the manipulated DLO.

Given this, it is crucial to take elastic instability into consideration when designing an optimal deployment scheme so that the robot's grasp and possible jittering of the manipulator does not introduce large undesired deformations of the DLO. 
To achieve this, we thoroughly analyze all potential robot grasps for manipulating a DLO in the $x$-$z$ plane to achieve a straight-line deployment. Our objective is to identify an optimal grasp that satisfies Eq.~\ref{eq::optimize} while effectively preventing the manipulated DLO from buckling due to elastic instability.

\subsubsection{Discovering Potential Grasp Region}
Given the suspended part's geometric properties and boundary conditions, we can write down the constraints $\mathcal{C}$ which should be satisfied:
\begin{equation}
\begin{aligned}
\bar{\mathbf q}(\bar s) \cdot \hat{\mathbf z} &\geq 0 \ \forall \ \bar s \in [0, \bar l_s], \\
\bar{\mathbf F}^\textrm{ext} \cdot \hat{\mathbf z} &\geq 0. \\
\end{aligned}
\label{eq:cons}
\end{equation}
These constraints enforce that (i) the suspended part should be above the substrate and (ii) external contact responses along the $z$-axis should always be larger than or equal to 0. 

By solving Eq.~\ref{eq::BVP} with constraints $\mathcal{C}$, we obtain all potential robot grasps of the manipulated end, forming a closed manifold $\mathcal {M}$ for a fixed normalized suspended length $\bar l_s$. 
The boundary condition at the connective node $\bar{\mathbf q}_C$ is defined as $\mathbf t(0) = (1, 0, 0)$ and $\bar \kappa= 0$.
Each point in the manifold $\mathcal{M}$ corresponds to a position $\bar{\mathbf q}_M$ and rotation $\mathbf R$ of the manipulated end. 
Given that the deformed configuration is located within the 2D $x$-$z$ plane, we can use a $2\times 1$ vector $\bar{\mathbf q}_M = (\bar x_\textrm{Top}, \bar z_\textrm{Top})$ to express the position of $\bar {\mathbf q}_M$ and a scalar value $\alpha$ to denote the rotation information.
For example, tangent $\mathbf t(\bar l_s) = (\cos(\alpha), \sin(\alpha))$ is shown in Fig.~\ref{fig::2Dschematic}(a). 
Since the manifold $\mathcal{M}$ is a closed set, we only need to obtain the boundary of the manifold $\partial \mathcal{M}$.

To discover the boundary $\partial \mathcal{M}$, we explore along a ray $\mathbf r$ from the connective node $\bar{\mathbf q}_C$ to the manipulated node $\bar{\mathbf q}_M$. 
The robot grasp along the ray can be divided into three regions as shown in Fig.~\ref{fig::2Dschematic}(b).
When the robot grasp exists in regions I and III, constraints $\mathcal C$ are not satisfied. In region I, the external force $\bar F_z = F_z h^2/k_b$ is smaller than 0, violating the constraints as stretching occurs, and in region III, the manipulated end is too low, leading to contact between the suspended part and the substrate.
Thus, region II, existing between regions I and III, represents the manifold $\mathcal{M}$ area that satisfies the constraints $\mathcal C$.
In this article, we implement a bisection method to obtain the boundary $\partial \mathcal{M}$ of region II. 
The pseudocode for the bisection method is given in Alg.~\ref{alg:bisectinon}.

\begin{figure}[t!]
\centerline{\includegraphics[width =\columnwidth]{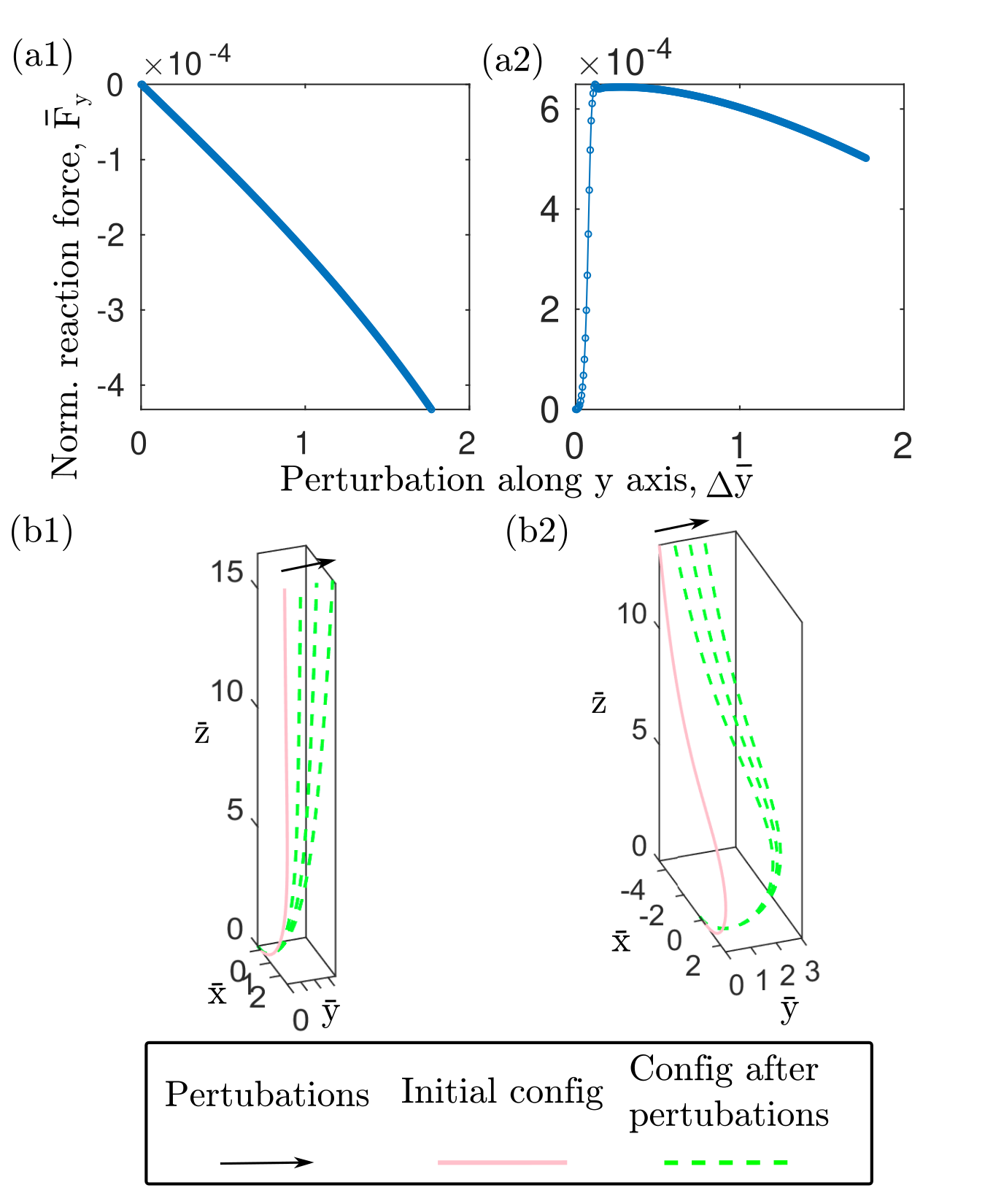}}
\caption{Change of the magnitude of normalized force $\bar F_y $ when adding a perturbation along the $y$-axis at local minimum 1 (a1) and local minimum 2 (a2), and change of the configurations of the rod when adding the perturbations at local minimum 1 (b1) and local minimum 2 (b2) for $\bar l_s = 17.68$.}
\label{fig::stability}
\end{figure}

\begin{algorithm}[h!]
\SetAlgoLined
\LinesNumbered
\DontPrintSemicolon
\KwIn{$\bar l_s, \bar k_s, \nu$}
\KwOut{$\partial \mathcal{M}$}
\SetKwProg{Fn}{Func}{:}{}
\SetKwFunction{DiscoverManifoldBoundary}{DiscoverManifoldBoundary}
\SetKwRepeat{Do}{do}{while}
{
\Fn{\DiscoverManifoldBoundary{$\bar l_s, \bar k_s$}}
{
$\theta \gets$ a small positive value \;
$\beta \gets$ a small positive value\;
$ \partial M \gets$ initialize an empty list \;
$\delta \gets$ a small positive value as tolerance \;
$\mathcal{R}\gets$ initialized rod solver with $\bar l_s, \bar k_s, \nu$ \;
\While{ $\theta \leq \pi$}{
$\mathbf r \gets (\bar l_s \cos(\theta), \bar l_s \sin(\theta))$ \;
\Do{$\bar F_z < 0$}
{$\mathbf r \gets (1 + \beta) \mathbf r$ \;
$ \bar F_z \gets \mathcal{R}(\mathbf r)$ \; }
$\mathbf r_c \gets \mathbf r$\;
\While{$\mathcal C$ \textup{is not satisfied}}{
$\mathbf r \gets \mathbf r - \delta \hat{\mathbf r}$\;
$\bar{\mathbf q}, \bar F_z \gets \mathcal{R}(\mathbf r)$\;
\uIf {$\lVert \mathbf r \rVert < 0$}
{break}
}
$\mathbf r_f \gets \mathbf r$ \;
\While{$\lVert \mathbf r_c - \mathbf r_f \rVert \geq \delta$}{
$\bar{\mathbf q}, \bar F_z \gets \mathcal{R}(\mathbf r)$\;
\uIf{$\mathcal{C}$ \textup{is satisfied}}
{$\mathbf r_f \gets \mathbf r$}\Else
{
$\mathbf r_c \gets \mathbf r$
}
$\mathbf r \gets (\mathbf r_c + \mathbf r_f)/2$\;
}
$\partial \mathcal{M}$.append($(\mathbf r \cos \theta, \mathbf r\sin\theta)$) \;
$\mathbf r_c \gets \mathbf r$\;
$\mathbf r_f \gets (0, 0)$\;
\While{$\lVert \mathbf r_c - \mathbf r_f \rVert \geq \delta$}{
$\bar{\mathbf q}, \bar F_z \gets \mathcal{R}(\mathbf r)$\;
\uIf{$\mathcal{C}$ \textup{is satisfied}}
{$\mathbf r_c \gets \mathbf r$}\Else
{
$\mathbf r_f \gets \mathbf r$
}
$\mathbf r \gets (\mathbf r_c + \mathbf r_f)/2$\;
}
$\partial \mathcal{M}$.append($(\mathbf r \cos \theta, \mathbf r\sin\theta)$) \;
$\theta \gets \theta + \delta \theta$\;
}
}
\textbf{return} $\partial \mathcal{M}$ \;
}
\caption{Bisection Method for Obtaining $\partial \mathcal{M}$}
\label{alg:bisectinon}
\end{algorithm}

Note that $\theta$ in Alg.~\ref{alg:bisectinon} is the angle between the $x$-axis and ray $\mathbf r$. 
A specific case for $\bar l_s = 17.68$ is visualized in Fig.~\ref{fig::2Dschematic}(c). 
Since deformations only occur in the $x$-$z$ plane, the twisting moment $\bar{\mathbf M}(0) = \mathbf M(0)h/k_b$ applied on the connective node $\bar{\mathbf q}_C$ is always 0.
To achieve the optimal pose of the manipulated end for $\bar l_s = 17.68$, we need to find the poses in $\mathcal{M}$ that minimizes $\lVert \bar{ \mathbf F}^\textrm{ext} \rVert$. 
Two local minima are found in the case shown in Fig.~\ref{fig::2Dschematic}(c), corresponding to two solutions of Eq.~\ref{eq::optimize}.  As stated before, we must select the local minima corresponding to the stable deformed suspended part.

\subsubsection{Checking Elastic Instability via Perturbations}
To test the elastic stability of these local minima, we apply a disturbance $\Delta{\bar y} = y/L_{gb}$ along the $y$-axis while the manipulated end $\bar{\mathbf q}_M$ is at each local minimum. 
Fig.~\ref{fig::stability} illustrates the changes in $\bar F_y = F_y h^2/k_b$ and the configurations resulting from these perturbations for each local optimum.

For local minimum 2, we can see a sudden snapping process, where an immediate change can be observed, while the disturbance on local minimum 1 results in a continuous and steady change.
Therefore, we can conclude that the optimum for deploying the DLO is at a local minimum 1 since this minimum corresponds to a configuration with more gentle bending deformations of the suspended part. 

Here, we also illustrate that the neighboring region around the elastic instability points has a higher tendency for significant deformations when the jittering of the manipulator occurs. In simulation, we introduce a small disturbance of $\Delta \bar{y} = 0.12$ along the $y$-axis for all potential robot grasps on the manifold $\mathcal{M}$. Fig.~\ref{fig::2Dschematic}(d) illustrates the maximum displacement of the suspended part along the y-axis $\Delta q_y^\textrm{max} = \max_{0 \leq s \leq l_s}(\mathbf q(s) \cdot \hat{\mathbf y})$ caused by this small disturbance. It is evident from the results that the neighboring region around local minimum 2 exhibits a higher tendency for significant deformations along the $y$-axis. Consequently, robot grasps within this region are more likely to induce instability in the manipulated DLO.

We can now output the optimal deployment rule for a straight line using the method introduced in this section. 
In the next section, we focus on optimal 3D manipulation, i.e., deploying patterns with nonzero curvature.
The following section discusses how to use a first-order optimization algorithm to solve Eq.~\ref{eq::optimize} for deploying any arbitrary prescribed pattern, where the optima for straight-line deployment is used as seeds when searching for the optima of more complex patterns.

\subsection{Deployment in 3D Workspace}

\begin{figure*}[t!]
\centerline{\includegraphics[width =\textwidth]{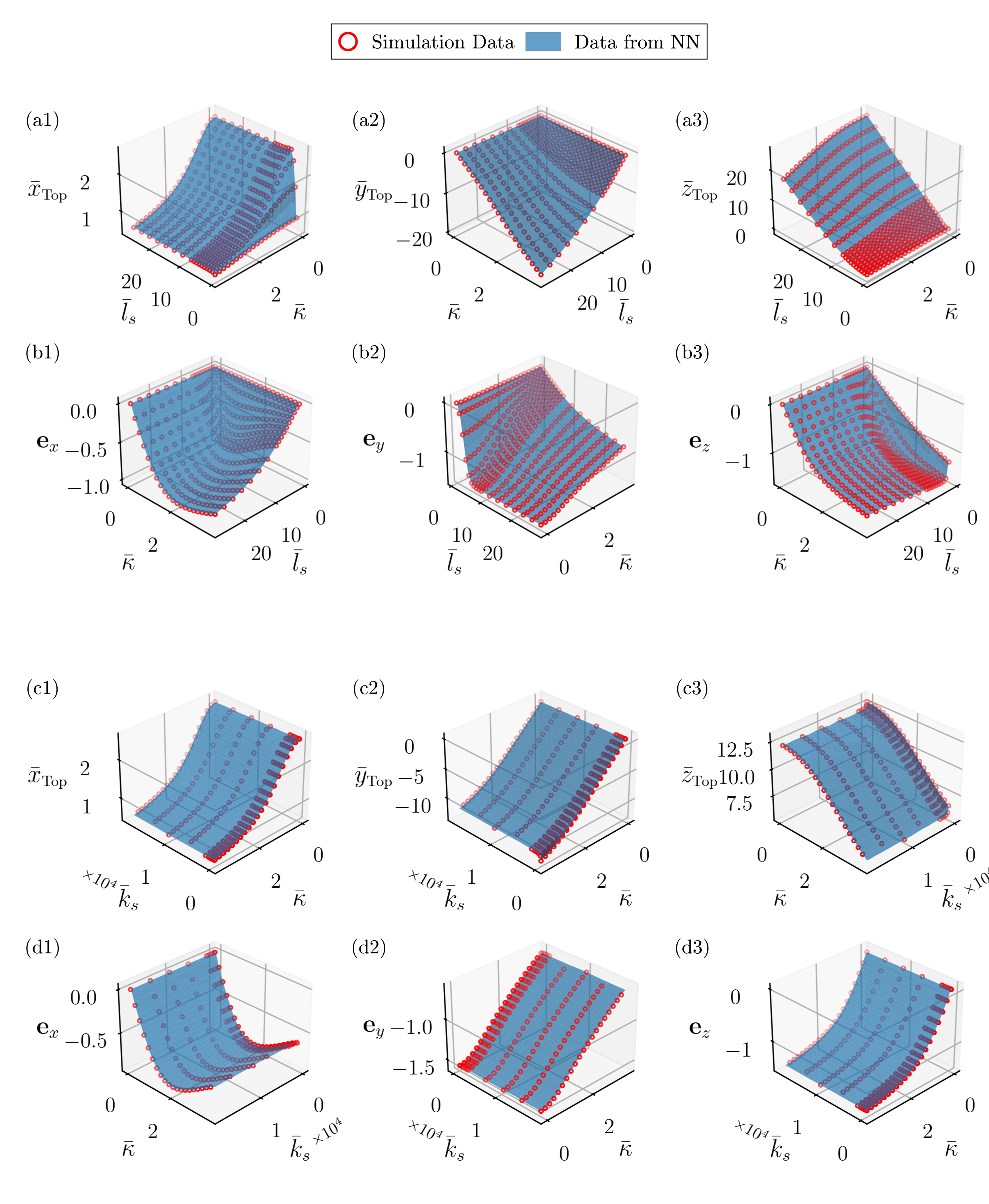}}
\caption{Visualization of the influence from curvature $\bar \kappa$ and suspended length $\bar l_s$ on the (a1-a3) manipulated end position and (b1-b3) manipulated end orientation for fixed values of $\bar k_s = 2087$; visualization of the influence from stretching stiffness $\bar k_s$ and curvature $\bar \kappa$ on the (c1-c3) manipulated end position and (d1-d3) manipulated end orientation with fixed values of $\bar l_s = 13.68$. }
\label{fig::3DDeployment}
\end{figure*}

As mentioned in Sec.~\ref{sec::physicalAnalysis}, the mapping relationship $\mathcal{F}(\cdot)$ in Eq.~\ref{eq::nondimensionl_mapping} must be constructed to achieve optimal deployment in the 3D workspace. 
For the connective node of any prescribed pattern, since the deformations of the pattern are only in the $x$-$y$ plane, we can ensure that the twisting moment $\mathbf M(0)$ can always be 0.
Therefore, the optimal pose of the manipulated end can be obtained by minimizing $\lVert \bar{\mathbf F}^\textrm{ext} \rVert$ by solving
\begin{equation}
    \grad_{\bar{\mathbf q}_M} \lVert \bar{\mathbf F}^\textrm{ext} \rVert = \frac{\partial \bar{\mathbf F}^\textrm{ext}}{\partial \bar{\mathbf q}_M} \bar{\mathbf F}^\textrm{ext} = 0.
\label{eq::3DGrad}
\end{equation}
As the deploying rod is a continuous system, $\bar{\mathbf F}^\textrm{ext}$ must change when $\mathbf q_M$ changes. 
Therefore, we can convert Eq.~\ref{eq::3DGrad} to be a root finding problem
\begin{equation}
     \bar{\mathbf F}^\textrm{ext} = 0.
\label{eq::3Doptimize}
\end{equation}

As discussed before, solving the configurations of the deploying DLO is a boundary value problem. 
Since the pattern's shape determines the boundary conditions on the connective end, the external forces $\bar{\mathbf F}^\textrm{ext}$ are influenced solely by the manipulated end pose $\mathbf q_M$, with a unique corresponding $\mathbf R$ for describing the rotation of the manipulated end.

Given the high nonlinearity of the DLO, it is nontrivial to solve the root-finding problem in Eq.~\ref{eq::3Doptimize} analytically. 
Therefore, we employ a finite difference approach to calculate the numerical Jacobian of $\mathbf F^\textrm{ext}$. We perturb the manipulated end along $x$, $y$, and $z$-axes with a small distance $\delta$ and use the finite difference to compute the numerical Jacobian

\begin{equation}
\mathbf J^\textrm{ext} = 
\frac{1}{\delta} \\
\begin{bmatrix}
\bar{\mathbf F}^\textrm{ext}(\bar{\mathbf q}_M + \delta \hat{\mathbf x}) - \bar{\mathbf F}^\textrm{ext}(\bar{\mathbf q}_M), \hfill \\
\bar{\mathbf F}^\textrm{ext}(\bar{\mathbf q}_M + \delta \hat{\mathbf y}) - \bar{\mathbf F}^\textrm{ext}(\bar{\mathbf q}_M), \hfill \\
\bar{\mathbf F}^\textrm{ext}(\bar{\mathbf q}_M + \delta \hat{\mathbf z}) - \bar{\mathbf F}^\textrm{ext}(\bar{\mathbf q}_M) \hfill
\label{eq::jacobian}
\end{bmatrix}^T,
\end{equation}
where $T$ is the transpose operator and $\delta \hat{\mathbf x}, \delta \hat{\mathbf y},$ and $\delta \hat{\mathbf z}$ are small perturbations along $x$, $y$, and $z$-axes, respectively, i.e., $\delta \hat{\mathbf x} = \left[ \delta, 0, 0 \right]^T$.

Here, $\mathbf J^\textrm{ext}$ is a $3 \times 3$ matrix and can be used to calculate the Newton search step so that Eq.~\ref{eq::3Doptimize} can be solved with a gradient descent method. 
Further details of this solving process are stated in Alg.~\ref{alg:gradient_descent}. 
Additionally, we also implement a line search algorithm to help determine the appropriate step size for the Newton search step $\Delta \bar{\mathbf q}$ as shown in Alg.~\ref{alg:line_search}.

\begin{algorithm}[!t]
\SetAlgoLined
\LinesNumbered
\DontPrintSemicolon
\KwIn{$\bar l_s, \bar \kappa \hat{\mathbf y}, \bar k_s, \nu$}
\KwOut{$\mathbf q_M^\star$}
\SetKwProg{Fn}{Func}{:}{}
\SetKwFunction{OptimalGrasp}{OptimalGrasp}
\SetKwFunction{LineSearch}{LineSearch}
\SetKwRepeat{Do}{do}{while}
{
\Fn{\OptimalGrasp{$\bar l_s, \bar \kappa, \bar k_s$}}
{
$k \gets 0$ \;
$\delta \gets$ a small value as tolerance \;
$ \bar{\mathbf q}_M^{(k)} \gets$ initialize a random pose of end-effector\;
$ \mathcal{R}(\cdot) \gets$ initialize the rod solver with $\bar l_s, \bar \kappa, \bar k_s$\;
\Do{$\lVert \bar{\mathbf F}^{\textrm{ext}} \rVert \geq \delta$}{
$\bar{\mathbf F}^{\textrm{ext}} \gets \mathcal{R} (\bar{\mathbf q}_M^{(k)})$\;
$\mathbf J^\textrm{ext} \gets$ Eq.~\ref{eq::jacobian} \;
$\Delta \bar{\mathbf q} \gets (\mathbf J^\textrm{ext})^{-1} \bar{\mathbf F}^{\textrm{ext}}$\;
$\alpha \gets$ \LineSearch{$\bar{\mathbf q}_M^{(k)}, \Delta \mathbf q, \lVert \bar{\mathbf F}^\textrm{ext} \rVert , \mathcal{R}$}\;
$\bar{\mathbf q}_M^{(k+1)} \gets \bar{\mathbf q}_M^{(k)} - \alpha \Delta \bar{\mathbf q}$\;
$k \gets k+1$
}
$\bar{\mathbf q}_M^* \gets \bar{\mathbf q}_M^{(k)}$ \;
\textbf{return} $\bar{\mathbf q}_M^*$ \;
}
}
\caption{Gradient Descent for Optimal Grasp
}
\label{alg:gradient_descent}
\end{algorithm}

\begin{algorithm}[!t]
\SetAlgoLined
\LinesNumbered
\DontPrintSemicolon
\KwIn{$\bar{\mathbf q}_M, \Delta \mathbf q, f_0, \mathcal{R}$}
\KwOut{$\alpha$}
\SetKwProg{Fn}{Func}{:}{}
\SetKwFunction{LineSearch}{LineSearch}
\SetKwRepeat{Do}{do}{while}
{
\Fn{\LineSearch{$\bar{\mathbf q}_M, \Delta \mathbf q, f_0, \mathcal{R}, \alpha_0 = 1, m = 0.5$}}
{
$\alpha \gets \alpha_0$ \;
$k \gets 0$ \;
success $\gets$ False \;
\Do{\textup{not success}}{
$\bar{\mathbf q}_M^{(k)} \gets \bar{\mathbf q}_M - \alpha \Delta \mathbf q$ \;
$\bar{\mathbf F}^\textrm{ext} \gets \mathcal{R}(\bar{\mathbf q}_M^{(k)})$ \;
$ f^{(k)} \gets  \lVert \bar{\mathbf F}^\textrm{ext} \rVert$\;
\uIf{$f^{(k)} \geq f_0$}{
$\alpha \gets m\alpha$\;
$k \gets k+1$\;
}\Else{success $\gets$ True}
}
\textbf{return} $\alpha$ \;
}
}
\caption{Line Search Algorithm}
\label{alg:line_search}
\end{algorithm}

In this article, both position $\bar{\mathbf q}_M$ and rotation $\mathbf e$ of the manipulated end are represented as $3 \times 1$ vectors: $\bar {\mathbf q}_M = (\bar x^\textrm{Top}, \bar y^\textrm{Top}, \bar z^\textrm{Top})$ and $\mathbf e = (e_x, e_y, e_z)$. 
The rotation vector $\mathbf e$ can be translated to a rotation matrix through an axis-angle representation $(\hat{\mathbf e}, \lVert \mathbf e \rVert)$, where $\lVert \mathbf e \rVert$ is the rotation angle along the rotation axis $\hat{\mathbf e} = \mathbf e/\lVert \mathbf e \rVert$. 
For an input tuple $(\bar l_s, \bar \kappa, \bar k_s)$, we can now solve for the optimal pose of the manipulated end $(\mathbf q_M^*, \mathbf e^*)$. 
Visualizations of the discretely solved optimal poses obtained from simulation are shown as red hollow circles in Fig.~\ref{fig::3DDeployment}. 

We now know how to obtain the optimal manipulation pose given the input $(\bar l_s, \bar \kappa, \bar k_s)$ with simulations. A numeric solver based on simulations for generating the optimal trajectory for various 
prescribed patterns is released (see~\ref{github}).
However, solving for the optimal poses with the numeric solver makes real-time manipulation infeasible as trajectory generation can take several hours.
Instead, the following section introduces using a neural network to learn the optimal controlling rule for fast real-time inference.

\subsection{Training the Neural Controller}
Rather than obtaining the optimal grasp through the numerical solver detailed in the previous section, we train a neural network to learn an analytical approximation of $\mathcal F(\cdot)$ similar to the approach in \citet{choi2023deep}.
We use a simple fully-connected feed-forward nonlinear regression network consisting of 4 hidden layers, each with 392 nodes, as the network architecture.
Aside from the output, each layer is followed by a rectified linear unit (ReLU) activation. 

We frame the neural controller to have an input $\mathbf i \in \mathbb R^3$ and an output $\mathbf o \in \mathbb R^6$, where the input consists of the three non-dimensional values $\bar l_s$, $\bar \kappa$, and $\bar k_s$ and the output consists of two concatenated $3 \times 1$ vectors: the optimal position $\bar {\mathbf q}_M^*$ and rotation $\mathbf e^*$ of the manipulated end.
Using our simulation framework, we construct a dataset $\mathcal D$ consisting of 6358 training samples.

When training the neural controller, we first preprocess all inputs $\mathbf i$ through the standardization
\[
\mathbf i' = \frac{\mathbf i - \bar{\mathbf i}_\mathcal D}{\boldsymbol{\sigma}_\mathcal D},
\]
where $\bar{\mathbf i}_\mathcal D$ and $\boldsymbol{\sigma}_\mathcal D$ are the mean and standard deviation of the input portion of the dataset $\mathcal D$.
Afterward, we use an initial 80-20 train-val split on the dataset $\mathcal D$ with a batch size of 128.
We use mean absolute error (MAE) as our loss and use a training strategy that alternates between stochastic gradient descent (SGD) and Adam whenever training stalls.
In addition, the batch size is gradually increased up to a max size of 2048, and the entire dataset is used to train the controller once MAE reaches $< 0.003$.
With this scheme, we achieve a final MAE of $< 0.0015$.
The neural network's approximation of $\mathcal F(\cdot)$ can be seen visualized in Fig.~\ref{fig::3DDeployment}.

\section{Robotic System}
\label{sec:pereception-system-and-robotic-system}
\subsection{Perception System}

\begin{figure}[t!]
\centerline{\includegraphics[width =\columnwidth]{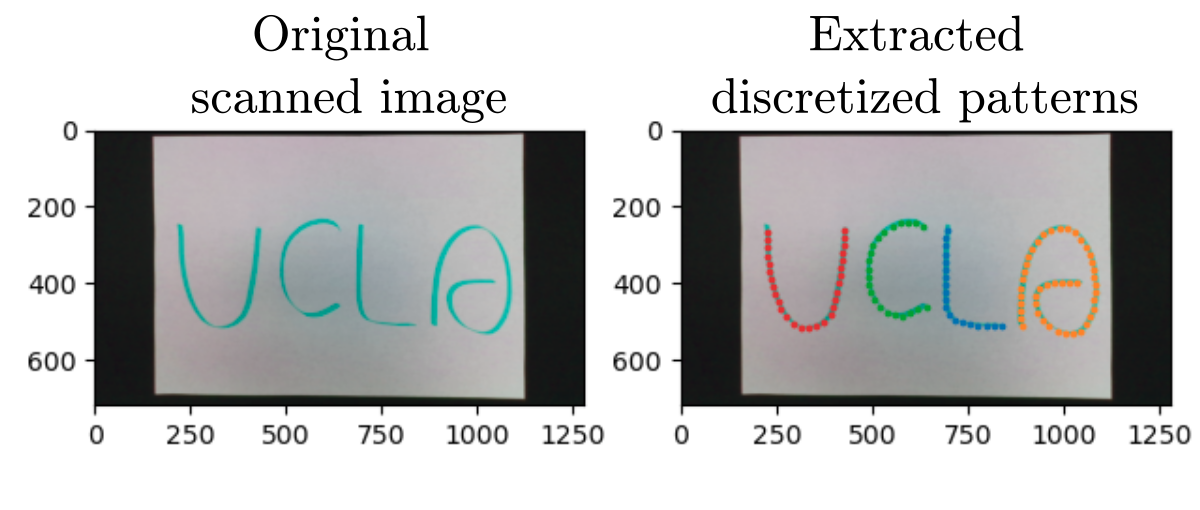}}
\caption{Handwritten letters and the corresponding extracted discretized patterns using mBEST~\citet{choi2023mBest}.}
\label{fig::scanned}
\end{figure}

To obtain the Cartesian centerline coordinates of the deployed DLO (or drawn patterns), we use the DLO perception algorithm mBEST~\citep{choi2023mBest}. 
This algorithm obtains the centerline coordinates of DLOs within an image by traversing their skeleton pixel representations.
The ambiguity of path traversal at intersections is handled by an optimization objective that minimizes the cumulative bending energy of the DLOs during the pixel traversal.
One case of extracting discretized patterns from the hand-writing pattern is shown in Fig.~\ref{fig::scanned}.
RGB images of the deployed DLO are obtained through an Intel RealSense D435 camera as shown in Fig.~\ref{fig:exp_apparatus}.
Further details of the perception algorithm itself can be found in the referenced paper.

\subsection{Motion Planning with the Neural Controller}
In Fig.~\ref{fig::first}, we showcase the full end-to-end pipeline of our proposed deployment scheme. 
Here, we give a full description of how to integrate the trained neural controller into a robot motion planner.

The first step of the deployment process is to specify the desired pattern.
This pattern can be defined by either an analytical function or detected as a drawn curve as shown in Fig.~\ref{fig::first}. 
Note that the pattern $\mathbf P(s)$ is treated as a function of the curve length $s$. 
Based on the configuration of the pattern, we can compute the required inputs for the neural controller when the connective node $\mathbf q_C$ achieves each point in the pattern $\mathbf P(s)$. 
The details of generating an optimal trajectory based on the pattern $\mathbf P(s)$ and the properties of the manipulated rod are given in Alg.~\ref{alg:opt}.

\begin{algorithm}[!t]
\SetAlgoLined
\LinesNumbered
\DontPrintSemicolon
\KwIn{$\mathbf P, L, L_{gb}, \bar k_s$ \\ Note that material parameters $L_{gb}, \bar k_s$ must be measured in advance (Fig.~\ref{fig:material} and Eq.~\ref{eq::material}).}
\KwOut{$\tau$}
\SetKwProg{Fn}{Func}{:}{}
\SetKwFunction{ProcessPattern}{ProcessPattern}
\SetKwFunction{DeploymentTrajectory}{OPT}
\SetKwFunction{AxangtoRot}{AxangtoRot}
\SetKwRepeat{Do}{do}{while}
{
\Fn{\DeploymentTrajectory{$L, \mathbf P, L_{gb}, \bar k_s$}}
{
$S, \kappa, \mathbf T \gets$ \ProcessPattern($\mathbf P$)  \;
$ \Delta s \gets$ step size of deployment \;
$\tau \gets$ initialize an empty list\;
$\hat{\mathbf z} \gets$ director along vertical direction \;
$s \gets$ 0 \;
\While{ $s \leq S$}{
$\mathbf q_C \gets \mathbf P(s)$ \;
$\hat{\mathbf x} \gets \mathbf T(s)$ \;
$\bar{\kappa} \gets \kappa(s) L_{gb}$\;
$\bar l_s \gets (L - s)/L_{gb}$ \;
$(\bar{\mathbf q}_M^*, \mathbf e^*) \gets \mathcal{F}(\bar l_s, \bar{\kappa}, \bar k_s) $ \;
$\mathbf R \gets$ \AxangtoRot$(\hat{\mathbf e}^*, \lVert \mathbf e^*\rVert)$\;
$ \mathbf R_t \gets (\hat {\mathbf x}, \hat {\mathbf z} \times \hat {\mathbf x}, \hat {\mathbf z}) $\;
$\mathbf q_M^* \gets \mathbf q_C + \mathbf R_t \bar{\mathbf q}_M^* L_{gb}$ \;
$\mathbf R^* \gets \mathbf R_t \mathbf R$ \;
Append $(\mathbf q_M^*, \mathbf R^*)$ to $\tau$ \;
$s \gets s + \Delta s$ \;
}
\textbf{return} $\tau$ \;
}
}
\caption{Optimal Deployment Trajectory}
\label{alg:opt}
\end{algorithm}

In Alg.~\ref{alg:opt}, $\kappa$ and $\mathbf T$ are all functions of the arc length $s$ of the pattern, where $\mathbf T$ is the tangent along the pattern. 
With Alg.~\ref{alg:opt}, we obtain the optimal grasp trajectory $\tau$ and then use OMPL~\citep{sucan2012open} to generate a valid motion planning sequence on a real robot system. 

One highlight of our overall robotic system is its real-time capability.
The real-time efficiency of the perception algorithm has been validated by~\citet{choi2023mBest} while the average end-to-end time to generate a full optimal deployment motion plan is less than 1 second. Therefore, our approach is also efficient enough for sensorimotor closed-loop control. 
However, as offline control has achieved excellent deployment accuracy in our experiments, online control is not carried out in this work. 

\section{Experiments and Analysis}
\label{sec:experiments}

\subsection{Measurement of Material Parameters}
\begin{figure}[t!]
\includegraphics[width =\columnwidth]{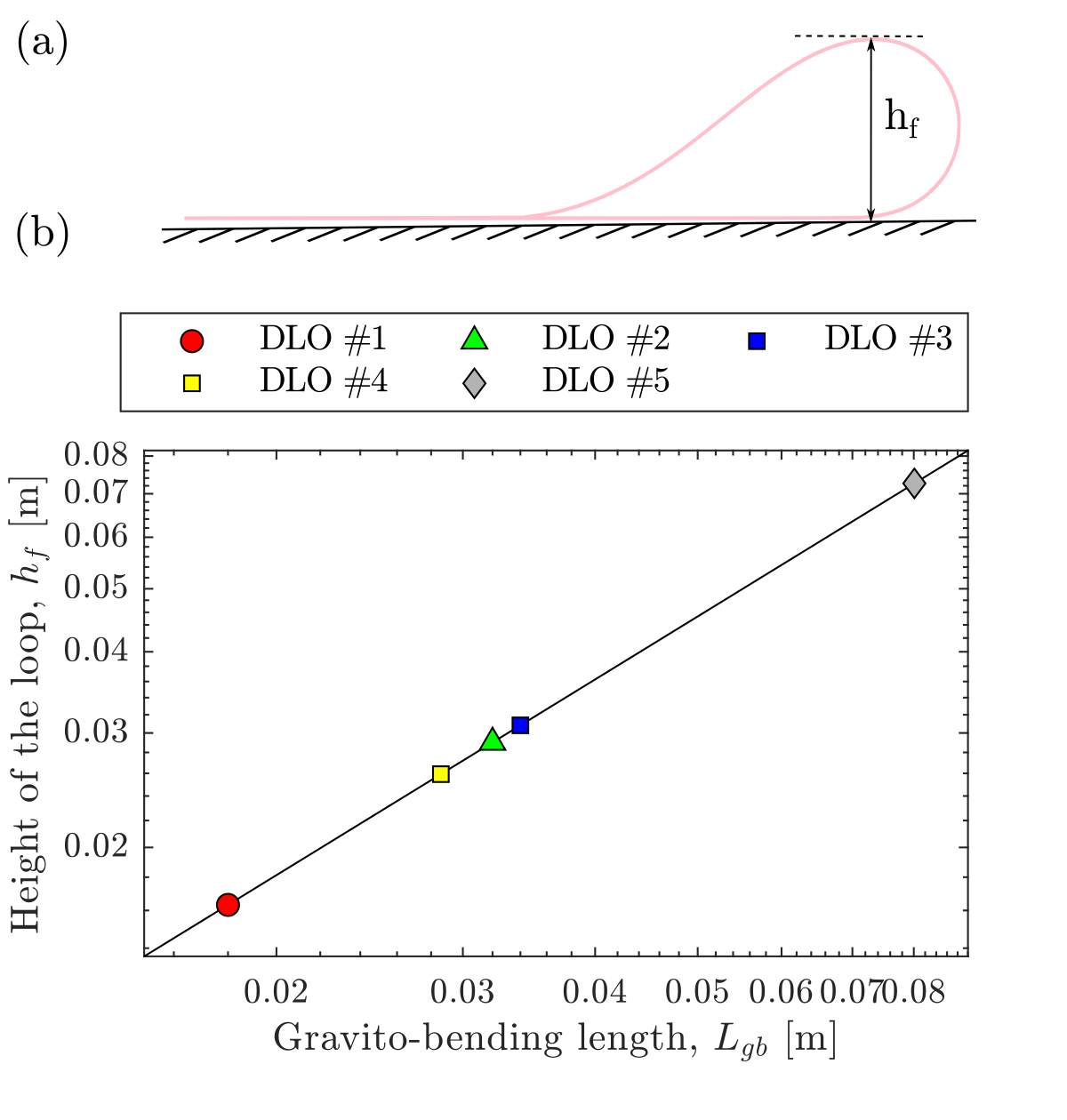}
\caption{(a) Deformed configurations of a DLO under gravity in 2D plane; (b) Relationship between the height of the loop $h_f$ and the gravito-bending length $L_{gb}$.}
\label{fig:material}
\end{figure}

To carry out deployment with our proposed scheme, we must validate its efficacy with comprehensive experiments. In this article, we choose to deploy various DLOs on different substrates for multiple tasks so that we can look into the robustness of the proposed scheme against the material difference and friction.

First, we need to find the geometric and material properties of the manipulated DLO. 
The geometry of the manipulated rod, e.g., total length $L$ and rod radius $h$, is trivial to measure. 
Measuring the material properties of the DLO is less clear.
Overall, we need to develop a way to find the following material properties: gravito-bending length $L_{gb}$ and normalized stretching stiffness $\bar k_s$.

Here, we presume the material is linearly elastic and incompressible. The incompressible material means the volume of the rod will not change when deformations happen. Therefore, Poisson's ratio is set as $\nu = 0.5$.
In addition, bending stiffness is $k_b = \frac{E\pi h^4}{4}$ where $E$ is Young's modulus, and the expression for gravito-bending length $L_{gb}$ and normalized stretching stiffness $\bar k_s$ is
\begin{equation}
\begin{aligned} 
L_{gb} &= \left(\frac{E h^2}{8 \rho g} \right)^{1/3}, \\
\bar k_s &= \frac{k_s L_{gb}^2}{k_b} = \frac{4 L_{gb}^2}{h^2}. \\
\end{aligned}
\label{eq::material}
\end{equation}

When observing Eq.~\ref{eq::material}, we find that the only parameter we must obtain is $L_{gb}$.
It is still unclear how to compute this as $L_{gb}$ is relevant to Young's Modulus $E$ and the density $\rho$ of the rod, which is usually hard to measure. 
Here, we propose a simple method that is able to measure $L_{gb}$ by observing the geometry of the rod. 
When we form a loop in a rod naturally using gravity in a 2D plane, we can observe the geometry of the rod becomes what is shown in Fig.~\ref{fig:material}(a). 
Indeed, the height $h_f$ of the loop has a linear relationship with $L_{gb}$. 
Therefore, we can obtain $L_{gb}$ for different rods by simply measuring $h_f$. 
According to prior work~\citep{pan2020periodic} and our validation shown in Fig.~\ref{fig:material}(b),
$h_f = 0.9066 L_{gb}.$

\begin{table*}[!t]
\renewcommand{\arraystretch}{1.1}
\centering
\caption{Material and geometric properties of the DLOs used in the experiments.}
\setlength\tabcolsep{0pt}
\begin{tabular*}{\textwidth}{@{\extracolsep{\fill}} ccccccccc}
\toprule
\multirow{2}{*}{DLO} & \multicolumn{7}{c}{Material \& Geometric Parameters} \\
\cmidrule(lr){2-9}
& Material & $L_{gb}$ [cm] & $h$ [mm] & $L$ [m] & $\nu$ & $\mu_\textrm{fabric}$ & $\mu_\textrm{steel}$ & $\mu_\textrm{foam}$ \\
\midrule
\#1 & Pink VPS & 1.8 & 1.6 & 0.875 & 0.5 & Low & Medium & High \\
\#2 & Green VPS & 3.2 & 1.6 & 0.885 & 0.5 & Low & Medium & High \\
\#3 & Rope & 3.4 & 2.0 & 0.89 & 0.5 & Medium & Low & High \\
\#4 & Pink VPS & 2.86 & 3.2 & 0.84 & 0.5 & Low & Medium & High \\
\#5 & Cable & 8.01 & 1.8 & 0.87 & 0.5 & Medium & Low & High \\
\bottomrule
\end{tabular*}
\label{tab:material}
\end{table*}

\begin{figure}[t!]
\centerline{\includegraphics[width =0.85\columnwidth]{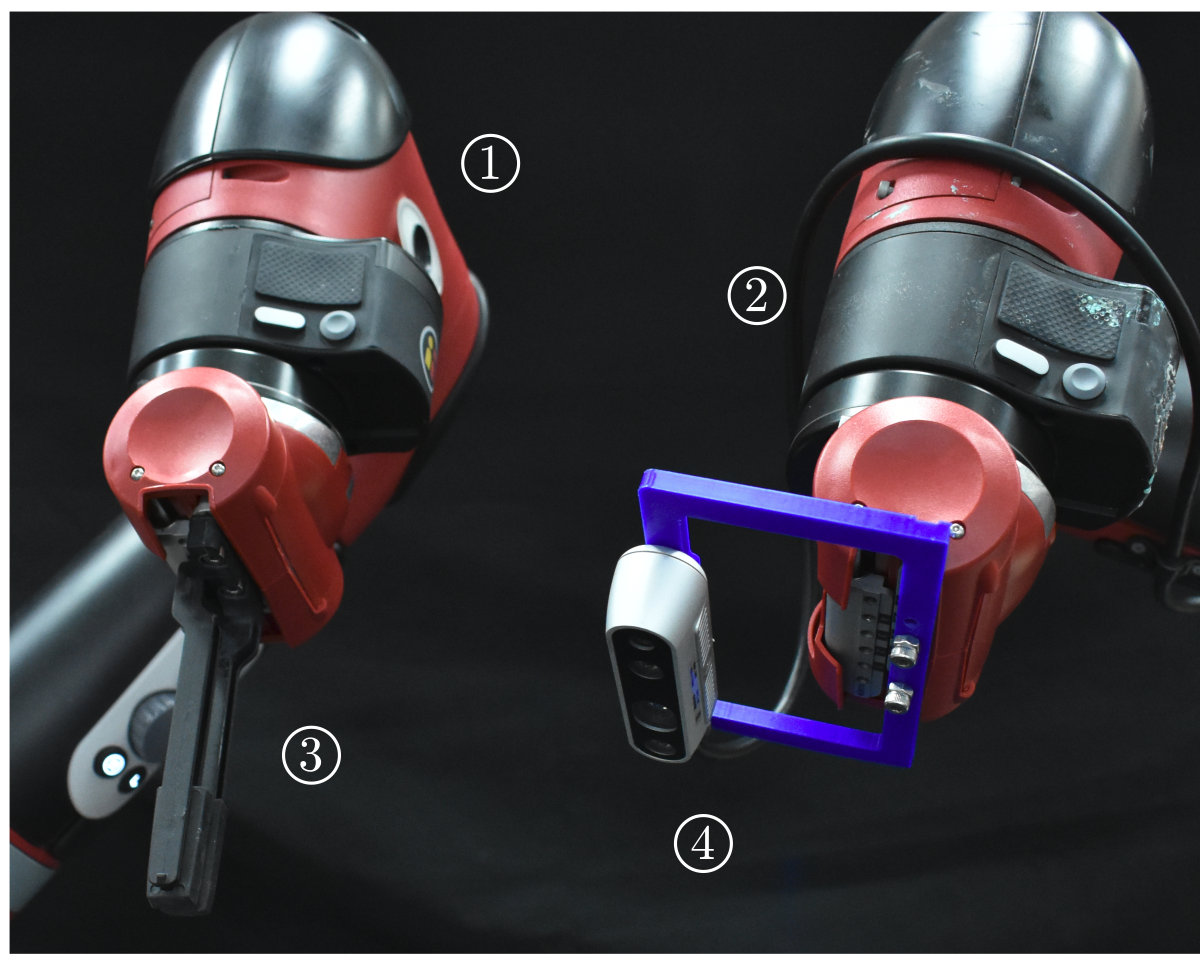}}
\caption{Experimental apparatus: Two robot manipulators, one for manipulation of the deploying rod (1) and the other for holding the camera for perception (2). A gripper (3) is used for grabbing the manipulated end of the rod. A camera (4) is used for extracting patterns from the drawn patterns and evaluating the deployment results.}
\label{fig:exp_apparatus}
\end{figure}

\subsection{Experiment Setup}
\label{sec:experimentSetup}

\subsubsection{Materials and Robot Hardware}
In this article, we conducted experiments involving five distinct types of DLOs. Among these, three are silicone-based rubber fabricated by vinyl polysiloxane (VPS); the fourth is a commercially available rope; and the fifth is a stiff USB cable. 
Note that we also validate the robustness of the deployment scheme against different substrates. 
The friction between the DLOs and substrates is also qualitatively measured.
Comprehensive details regarding the parameters for each of these DLOs can be found in Table~\ref{tab:material}.

For our experiments, we used two Rethink Robotics' Sawyer manipulators as shown in Fig.~\ref{fig:exp_apparatus}. 
One arm is attached with a gripper for manipulating the rod. 
The other arm holds an Intel RealSense D435 camera which is used to scan drawn patterns as well as obtain a top-down view of the deployment result for evaluations. 
A workstation with an AMD Ryzen 7 3700X CPU and an NVIDIA RTX 2070 SUPER GPU was used for all experiments.

\subsubsection{Experiment Tasks}
We implement our proposed deployment scheme across four distinct tasks. 
First, we deploy a rod along some canonical cases obtainable through analytical expressions such as a line, circle, and sine curve. 
The rod is deployed using the robotic arm with the gripper.
Once the deployment is finished, the other arm with the camera moves to scan the deployment result. 

The second task involves deploying patterns drawn on paper. Users draw patterns, subsequently scanned by the camera to obtain ordered discretized pattern coordinates. The robot then manipulates the rod to replicate the drawn pattern. This article showcases deployment results for the letters ``U'', ``C'', ``L'', and ``A'' with the precise shapes detailed in Fig.~\ref{fig::expVisualizedResult}(a).
The third task is geared towards validating the deployment scheme's application in cable placement, a vital aspect of cable management. The scheme's efficacy is demonstrated by placing cables along constrained paths with the help of pre-installed fixtures on the substrate.
Lastly, the deployment scheme's application for tying knots is verified. Both robotic arms are equipped with grippers for this task. 

For the first two tasks, patterns are evaluated using both intuitive and optimal control methods. Additionally, three different rods (DLOs $\#1$, $\#2$, and $\#3$) are deployed on substrates of various materials (fabric, steel, foam) to assess the method's robustness against material disparities and friction.
In the third task, DLO $\#5$ (USB cable) is employed for cable placement using both algorithms. 
Finally, DLOs $\#2$ and $\#4$ are used to tie distinct knots for the fourth task. Each experimental case is subjected to ten trials for each control method, culminating in a total of 1340 experimental trials. 

\subsection{Metrics}
We now formulate the metrics used to evaluate the performance of the deployment scheme. 
When deploying a pattern $\mathbf P$, we need to assess the accuracy of the deployment result. 
We first discretize the pattern $\mathbf P$ into $N$ points and denote the $i$-th point of the prescribed pattern as $\mathbf P^i$. 
The actual deployment pattern obtained from perception is denoted as $\mathbf P_{\textrm{exp}}$. 
With this discretization scheme, we compute the average error $e_\textrm{mean}$ and standard deviation $\sigma$ as
\begin{equation}
\begin{aligned} 
e_\textrm{mean} &= \frac{1}{N}\sum_{i=1}^{N}\lVert \mathbf P_\textrm{exp}^i - \mathbf P^i \rVert, \\
\sigma &= \sqrt{\frac{\sum_{i=1}^{N}(\lVert \mathbf P_\textrm{exp}^i - \mathbf P^i \rVert - e_\textrm{mean})}{N}}, \\
\end{aligned}
\label{eq::metric}
\end{equation}
for both the intuitive and optimal control results.

The accuracy evaluation is not applicable for the two application tasks: cable placement and knot tying as they are high-level tasks.
Therefore, we simply use the success rate of those application tasks to evaluate the performance of the deployment scheme. 
In addition to accuracy, we also report a detailed comparison of runtimes and errors between the numerical and NN-based solvers. 
Details of the relevant results and analysis are discussed in the next section.

\subsection{Results and Analysis}

\begin{figure*}[t!]
\centerline{\includegraphics[width =\textwidth]{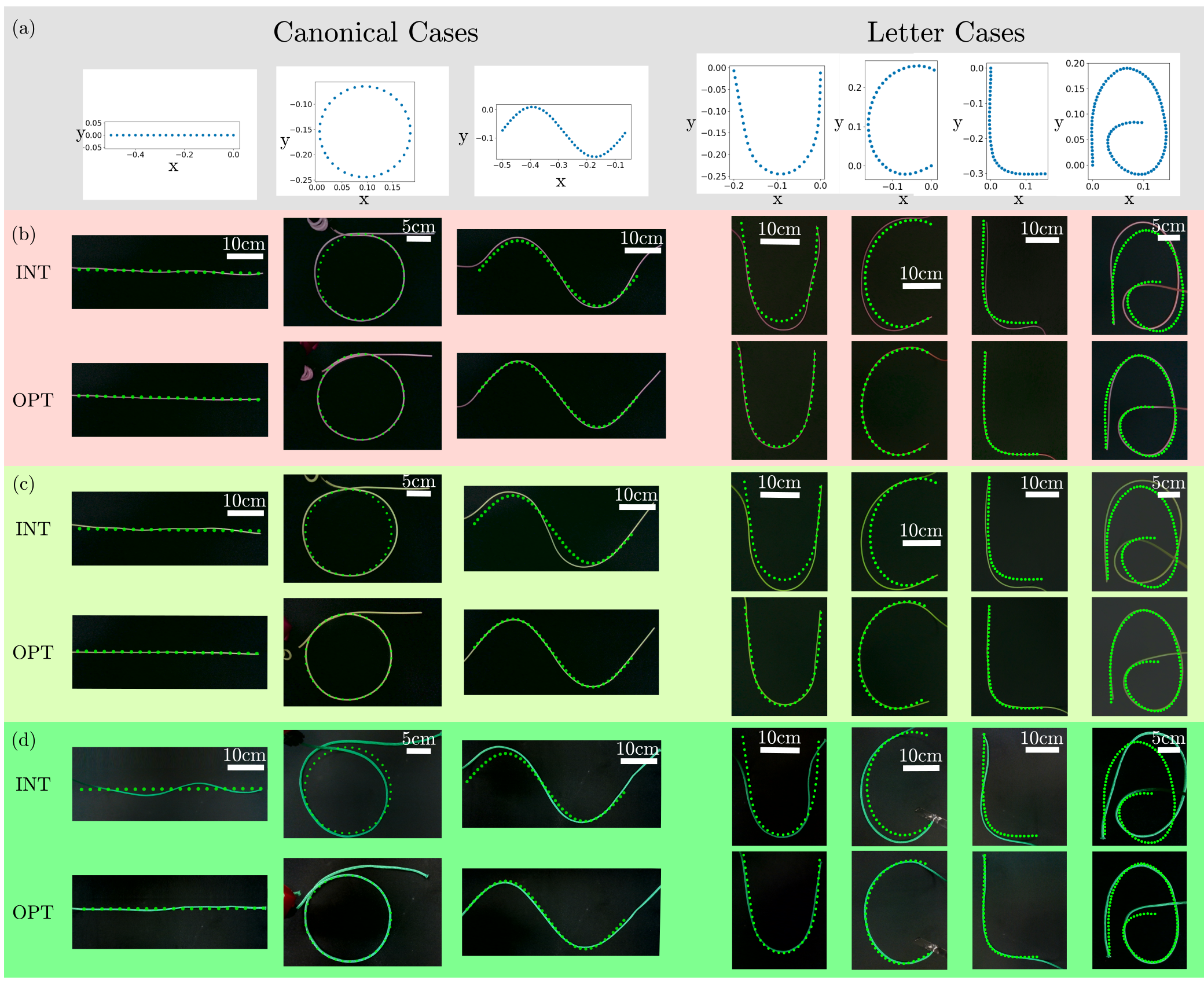}}
\caption{Experiment results of deployment along various patterns. (a) All used prescribed patterns are discretized and plotted. Deployment results for (b) DLO $\# 1$ (pink VPS), (c) DLO $\# 2$ (green VPS), and (3) DLO $\#3$ (rope) are shown for each prescribed pattern. Results for the intuitive control method and optimal control method are shown for each rod.} 
\label{fig::expVisualizedResult}
\end{figure*}

\begin{figure*}[t!]
\centerline{\includegraphics[width =\textwidth]{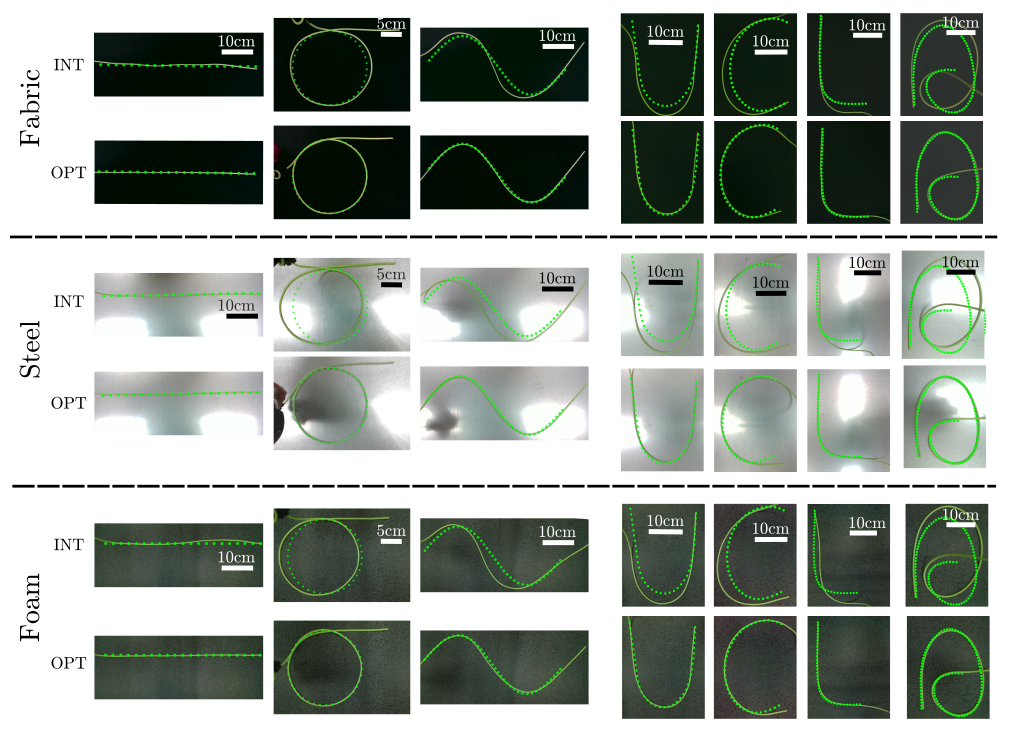}}
\caption{Experiment results of deployment with DLO $\#2$ (green VPS) along various patterns on different substrates.} 
\label{fig::greenVPSExp}
\end{figure*}

\begin{table*}[!t]
\renewcommand{\arraystretch}{1.1}
\centering
\caption{Evaluation of deployment accuracy for various patterns, DLOs, and substrates.}
\setlength\tabcolsep{0pt}
\begin{tabular*}{\textwidth}{@{\extracolsep{\fill}} llcccccccc}
\toprule
\multirow{2}{*}{DLO} & \multirow{2}{*}{SUB} & \multirow{2}{*}{\shortstack{Control \\ Scheme}} & \multicolumn{7}{c}{Pattern Type and Accuracy $e_\textrm{mean} \pm \sigma$ [cm] (Eq.~\ref{eq::metric})} \\
\cmidrule(lr){4-10}
& & & Line & Circle & Sine curve & Letter ``U'' & Letter ``C'' & Letter ``L'' & Letter ``A'' \\
\midrule
\multirow{6}{*}{\textrm{\#1}}
& \multirow{ 2}{*}{\textrm{Fabric}} 
& INT & $0.40\pm0.22$ & $0.61\pm0.36$ & $1.66\pm0.74$ & $1.39\pm0.63$ & $2.21\pm0.92$ & $1.00\pm0.59$ & $4.81\pm2.27$ \\
& & OPT & $0.14\pm0.09$ & $0.15\pm0.07$ & $0.27\pm0.10$ & $0.22\pm0.07$ & $0.22\pm0.10$ & $0.35\pm0.18$ & $0.47\pm0.23$ \\
& \multirow{ 2}{*}{\textrm{Steel}} 
& INT & $1.42\pm0.66$ & $2.34\pm1.24$ & $2.69\pm1.69$  & $3.59\pm2.39$& $3.67\pm1.93$ & $0.87\pm0.55$ & $3.64\pm2.09$ \\
& & OPT & $0.22\pm0.12$  & $0.22\pm0.08$ & $0.27\pm0.10$ & $0.24\pm0.13$ & $0.27\pm0.09$ & $0.42\pm0.16$ & $0.58\pm0.37$ \\
& \multirow{ 2}{*}{\textrm{Foam}} 
& INT & $1.03\pm0.21$ & $1.23\pm0.45$ & $2.84\pm1.52$ & $3.33\pm1.93$ & $3.89\pm1.29$ & $1.13\pm0.74$ & $4.09\pm2.19$ \\
& & OPT & $0.25\pm0.15$ & $0.18\pm0.06$ & $0.29\pm0.16$ & $0.24\pm0.15$ & $0.41\pm0.20$ & $0.35\pm0.12$ & $0.54\pm0.24$ \\
\midrule
\multirow{6}{*}{\textrm{\#2}}
& \multirow{ 2}{*}{\textrm{Fabric}} 
& INT & $0.52\pm0.13$& $1.64\pm0.95$ & $1.60\pm0.83$ & $3.74\pm2.89$ & $ 4.58\pm1.15$& $1.74\pm1.11$ & $4.95\pm2.63$ \\
& & OPT & $0.13\pm0.07$ & $0.16\pm0.07$ & $0.20\pm0.09$ & $0.17\pm0.11$  & $0.19\pm0.22$  & $0.29\pm0.11$ & $0.32\pm0.18$\\
& \multirow{ 2}{*}{\textrm{Steel}} 
& INT & $1.72\pm0.63$& $2.52\pm1.02$ & $3.30\pm2.08$ & $4.78\pm4.15$ & $6.66\pm2.53$ & $2.14\pm1.26$ & $5.23\pm3.38$ \\
& & OPT & $0.17\pm0.08$& $0.22\pm0.09$ &$0.54\pm0.20$ & $0.21\pm0.09$ & $0.74\pm0.31$& $0.66\pm0.24$ & $0.36\pm0.17$ \\
& \multirow{ 2}{*}{\textrm{Foam}} 
& INT &$1.38\pm0.60$ & $2.24\pm0.97$ & $4.17\pm2.57$ & $5.42\pm4.47$ & $6.14\pm3.08$ & $1.70\pm1.32$ & $5.09\pm3.39$ \\
& & OPT &$0.27\pm0.13$& $0.20\pm0.09$ & $0.37\pm0.14$& $0.17\pm0.08$  & $0.39\pm0.18$  & $0.37\pm0.15$ & $0.43\pm0.19$ \\
\midrule
\multirow{6}{*}{\textrm{\#3}}
& \multirow{ 2}{*}{\textrm{Fabric}} 
& INT & $1.56\pm0.81$ & $1.13\pm0.53$ & $5.09\pm1.35$ & $4.22\pm3.10$ & $3.36\pm1.58$ &$2.37\pm1.56$ & $4.59\pm2.54$\\
& & OPT & $0.49\pm0.28$ & $0.29\pm0.15$ & $0.47\pm0.23$ & $0.36\pm0.18$  & $0.35\pm0.19$ & $0.50\pm0.24$ & $0.56\pm0.29$ \\
& \multirow{ 2}{*}{\textrm{Steel}} 
& INT & $4.53\pm2.80$ & $1.85\pm0.45$ & $4.43\pm2.82$ & $4.53\pm2.80$& $3.35\pm1.55$ & $2.57\pm1.62$ & $4.30\pm1.73$ \\
& & OPT & $0.47\pm0.20$ & $0.29\pm0.13$  & $0.46\pm0.20$ & $0.47\pm0.20$ & $0.56\pm0.20$ &$0.51\pm0.24$ & $0.81\pm0.30$ \\
& \multirow{ 2}{*}{\textrm{Foam}} 
& INT &$2.00\pm0.88$ & $1.94\pm0.84$ & $3.80\pm1.96$ & $3.67\pm2.46$ & $6.03\pm3.11$ & $3.32\pm1.80$ & $4.47\pm2.50$\\
& & OPT & $0.78\pm0.34$ & $0.27\pm0.15$ & $0.46\pm0.20$ & $ 0.32\pm0.16$ & $0.56\pm0.26$ & $0.33\pm0.14$ &$0.52\pm0.20$\\
\bottomrule
\\
\end{tabular*}
\label{tab:results}
\end{table*}

\subsubsection{Accuracy}
All experimental results can be seen in Table~\ref{tab:results}.
To compute the error metrics in Eq.~\ref{eq::metric}, we used a discretization of $N = 50$. From all results, we can observe a noticeable improvement in our optimal control method over the intuitive method for various geometrical, material, and environmental parameters.

To better visualize our method's generality, we visually depict deployment outcomes across different DLOs on the fabric surface in Fig.~\ref{fig::expVisualizedResult}. In addition, a comparative visual representation of deployment results for a single DLO (\#2) on varying substrates is shown in Fig.~\ref{fig::greenVPSExp}. Readers seeking comprehensive visual comparisons of all deployment outcomes can refer to the supplementary video for detailed insights (see Footnote~\ref{github}).

Among the seven deployed patterns, the first three (straight line, circle, and sine curve) are canonical cases, i.e., their shapes have explicit analytical expressions. 
Note that when deploying the circle and sine curve patterns, a small ``remainder'' section is first deployed. 
This is necessary as the circle and sine curve patterns have a non-zero curvature at the start of their pattern.
We compensate for this by deploying a remainder part whose curvature gradually evolves from a straight line with 0 curvature to the prescribed curvature of the pattern's first point. The remainder can improve the deployment task's accuracy as the deployed pattern will require slight friction based on Eq.~\ref{eq::bendingFriction}.

We have omitted the designed remainder for the four remaining patterns denoted by the letters ``U'', ``C'', ``L'', and ``A'' for better visualization. Among these, patterns ``U'', ``L'', and ``A'' exhibit a relatively low $\kappa''$ value during the beginning stage of the deployment, resulting in the deployment accuracy being minimally affected by surface friction. 

Conversely, the ``C'' pattern demonstrates a comparatively higher $\kappa''$ value initially, leading to a possible noticeable mismatch between the deployed DLO and the intended pattern in the beginning.
The impact of friction becomes more pronounced during the rope deployment corresponding to DLO $\#3$ since the rope has higher bending stiffness $k_b$ and experiences lower friction with the substrate. Fixing the free end is essential to precisely replicate the ``C'' pattern with the rope as shown in Figure~\ref{fig::expVisualizedResult} (d). Despite this limitation, our optimized deployment strategy consistently outperforms the intuitive approach.

\subsubsection{Computational Efficiency}
Next, we also evaluated the computational efficiency of our neural controller. Table~\ref{tab:timeDiff} compares time costs between the neural network solver (NN-solver) and the numeric solver based on simulations. 
When calculating a single optimal robot grasp for a given parameter tuple ($\bar l_s$, $\bar \kappa$, $\bar k_s$), the numeric solver takes approximately 10 to 20 seconds, while our NN-solver takes roughly 0.4 seconds.

The difference of time costs becomes more significant when generating a series of optimal robot grasps for a discretized pattern. 
Note that a discretized pattern typically consists of 100 to 200 nodes and that the numeric solver needs to compute the robot trajectory in sequence as the optimal grasp for the previous step is needed as the seed for computing the next optimal grasp. 
Therefore, the time costs quickly accumulate for the numeric solver, which substantially elongates the overall computation time. 
In contrast, the NN-solver leverages vectorization to solve multiple robot grasps simultaneously, resulting in a speed advantage of several orders of magnitude compared to the numeric solver when generating optimal deployment trajectories. 

\subsubsection{Precision of the Neural Controller}
Finally, Table~\ref{tab:timeDiff} also presents the precision of the NN-solver. The solutions from the numeric solver serve as the ground truth. Mean Absolute Error (MAE) is employed to evaluate the optimal trajectories the NN-solver generates against the ground truth. Remarkably, the MAE consistently remains below $0.003$m for position error and $0.009$ for differences in rotation quaternions. Importantly, it's noteworthy that none of the solved trajectories in this analysis were part of the training dataset. Thus, we can confidently assert that our NN-solver exhibits robustness, efficiency, and accuracy, rendering it well-suited for real-time control applications.

\begin{table*}[!t]
\renewcommand{\arraystretch}{1.1}
\centering
\caption{Evaluation of computation times of various patterns for the numerical and NN-solvers with error metrics.}
\setlength\tabcolsep{0pt}
\begin{tabular*}{\textwidth}{@{\extracolsep{\fill}} llcccccccc}
\toprule
\multirow{3}{*}{\textrm{DLO}} & \multirow{3}{*}{\shortstack{Solver Times [s] \\ \& MAEs}} & \multicolumn{7}{c}{Patterns with Number of Nodes} \\
\cmidrule(lr){3-9}
& 
 &  Line & Circle & Sine curve & Letter ``U'' & Letter ``C'' & Letter ``L'' & Letter ``A'' \\
 & & 101 nodes & 156 nodes & 138 nodes & 190 nodes & 190 nodes & 190 nodes & 194 nodes \\
\midrule
\multirow{4}{*}{\textrm{\#1}}
& Numeric-Solver &   $1572.68$ &  $2036.11$ & $2897.17$ & $3954.12$ & $4015.24$ & $4777.30$ & $4666.55$\\
& NN-Solver & $0.402$& $0.393$& $0.395$&  $0.431$&$0.431$ &$0.400$ &$0.417$\\
& Position Error [m] & $0.0008$ & $0.0007$& $0.0009$& $0.0008$& $0.0007$ & $0.0008$ & $0.0008$ \\
& Orientation Error & $0.0012$ & $0.0010$& $0.0032$& $0.0025$& $0.0020$ & $0.0020$ & $0.0021$ \\
\midrule
\multirow{4}{*}{\textrm{\#2}}
& Numeric-Solver & $776.56$  &$1213.14$  &$1769.66$ & $2286.66$ & $2226.73$ & $2720.08$ & $2933.90$\\
& NN-Solver & $0.397$&  $0.391$ &  $0.396$ &$0.419$ & $0.408$ & $0.404$ &  $0.406$ \\
& Position Error [m] & $0.0016$ & $0.0016$& $0.0019$& $0.0018$& $0.0016$ & $0.0020$ & $0.0017$ \\
& Orientation Error & $0.0012$ & $0.0078$& $0.0050$& $0.0042$& $0.0020$ & $0.0058$ & $0.0030$ \\
\midrule
\multirow{4}{*}{\textrm{\#3}}
& Numeric-Solver & $666.01$ & $1041.71$&$1561.12$ &$1984.63$  &  $1972.39$ &  $2405.71$&  $2639.44$\\
& NN-Solver &$0.400$ &$0.407$ & $0.395$ & $0.407$  &  $0.420$& $0.405$&  $0.411$\\
& Position Error [m] & $0.0016$ & $0.0017$& $0.0020$& $0.0020$& $0.0016$ &$0.0021$ & $0.0018$ \\
& Orientation Error & $0.0010$ & $0.0087$& $0.0052$& $0.0055$& $0.0023$ &$0.0054$ & $0.0032$ \\
\bottomrule
\end{tabular*}
\label{tab:timeDiff}
\end{table*}

\subsection{Application \#1: Cable Placement}
\begin{figure}[t!]
\centerline{\includegraphics[width =\columnwidth]{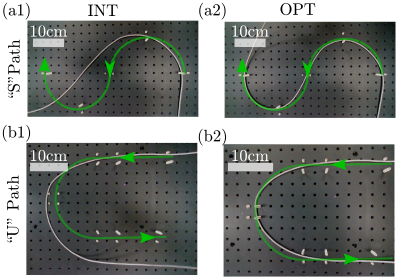}}
\caption{A demonstration of cable placement along different prescribed patterns with both intuitive and optimal control schemes.}
\label{fig::cable}
\end{figure}
In this section, we showcase the application of the deployment scheme for cable placement. The importance of cable management has surged, particularly in engineering contexts involving tasks like wire harnessing, infrastructure development, and office organization~\citep{sanchez2018robotic, lattanzi2017review}. Given cables' inherent high bending stiffness, shaping them to specific forms can be challenging, often necessitating external fixtures to maintain the desired configuration. When humans perform cable management manually, meticulous placement along the designated pattern is essential, coupled with the use of fixtures to secure the cable in place. However, a robotic system can autonomously execute cable placement with our designed optimal deployment strategy.

In our experimental setup, we preinstalled external fixtures into the stainless steel breadboard to delineate the intended patterns. These fixtures also counteract the cable's rigid nature, preventing it from reverting to its original shape. The deployment results can be visualized in Figure~\ref{fig::cable}.  Compared to the failure placement results with the intuitive scheme, our optimal deployment scheme can place the cable along the prescribed pattern ``U'' and ``S'' on the substrate.  We did 10 experimental trials for each deployment task illustrated in Fig.~\ref{fig::cable}. Notably, the optimal deployment approach achieved an impressive 90\% (9/10) success rate for both patterns, whereas the intuitive method failed in all trials (0/10) as shown in Table~\ref{tab:knot}.

\subsection{Application \#2: Knot Tying}
\begin{figure*}[t!]
\centerline{\includegraphics[width =\textwidth]{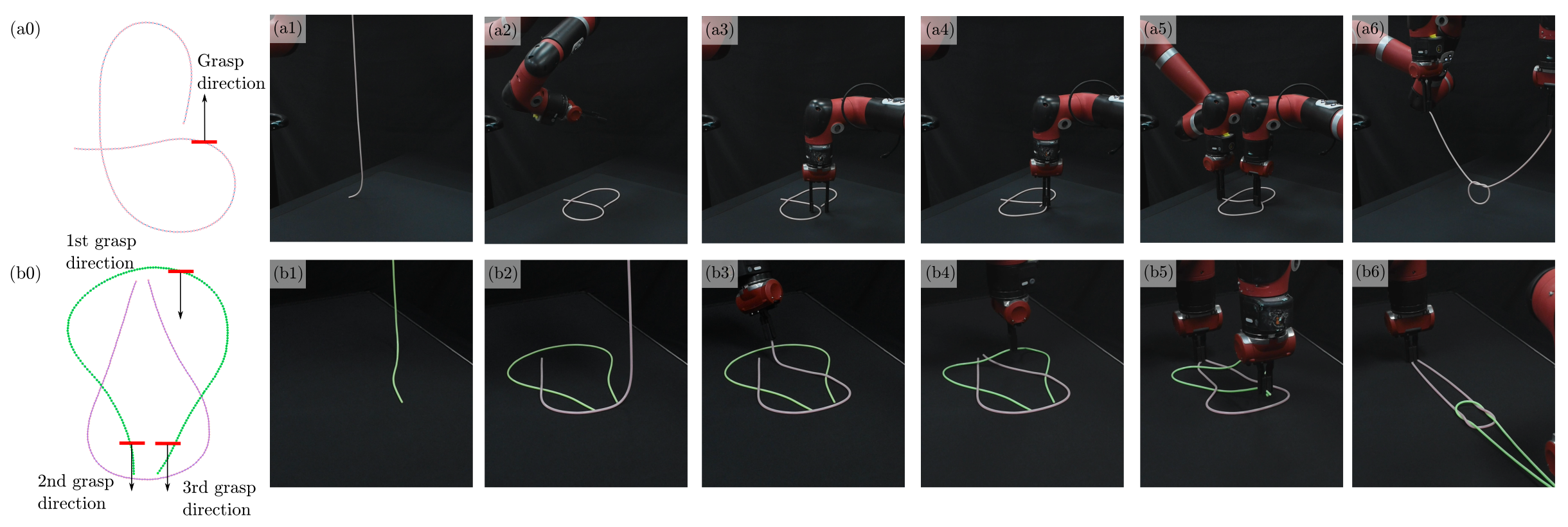}}
\caption{A demonstration of two knot-tying cases using the DLO deployment scheme. (a0) and (b0) are designed patterns for the trefoil knot and the reef knot, respectively. Time marches for trefoil knot from (a1) to (a6) and reef knot from (b1) to (b6).}
\label{fig::knot}
\end{figure*}

Since our optimal deployment scheme can control the shape of various DLOs with excellent accuracy, we can use this scheme to tie knots. 
First, the manipulated rod is deployed along a predesigned pattern on the substrate. 
Users can draw the predesigned pattern so that only a few extra manipulations are required.
Then, the camera will scan the drawn pattern and send it as input to our designed scheme.
The deployed pattern is designed in a way that only a few simple pick-and-place operations on certain knot segments is required to complete the tying sequence. 
Since the prescribed pattern's shape is known in advance, we can let the robot execute the pick-and-place procedure without perception feedback. 
So long as the initial deployment is accurate and repeatable, the subsequent pick-and-place procedure should succeed most of the time.

We showcase two knot-tying sequences in Fig.~\ref{fig::knot}. The top row showcases a trefoil knot, one of the most fundamental knots in engineering~\citep{crowell2012introduction}. 
For this knot, we used DLO $\#4$.
Another case is a reef knot, a prevalent knot widely used in for various applications including shoelaces, packaging, sewing, etc. When tying the reef knot, we used DLOs $\#2$ and $\#4$. Although these two DLOs have totally different material properties, our generalizable neural controller allows two robots to deploy both DLOs accurately along the designed patterns. 
With the help of the deployed patterns, reef knots can be tied with simple pick-and-place procedures.
Such knot-tying cases strongly support the potential of our deployment scheme in various engineering applications.

We show the results of the knot-tying tasks in Table~\ref{tab:knot}. The successful rate of knot-tying is remarkable. We achieve a success rate of 90 \% (9 successful trials out of 10) for tying a trefoil knot and a success rate of 70 \% (7 successful trials out of 10) with the optimal control method. Based on our observations, all the failure cases were caused by the rod slipping out of the gripper. In contrast, the intuitive control method achieves a success rate of 0\% for both cases as the initially deployed pattern does not match the intended pattern.

Therefore, the intuitive control method would require some visual feedback to choose the pick-and-place motion adaptively for the trefoil knot case. As for the reef knot case, due to the deployment results are totally wrong, even though the visual feedback is applied, it is still hard to achieve a complete reef knot with intuitive method.

\begin{table}[!t]
\renewcommand{\arraystretch}{1.1}
\caption{Real-world application experiment results.}
\centering
\begin{tabular}{lcc}
\toprule
Experiment Type & Scheme & Success rate \\
\midrule
\multirow{2}{*}{\textrm{``S'' Cable Placement}}
& INT & 0/10 \\
& OPT & 9/10 \\
\midrule
\multirow{2}{*}{\textrm{``U'' Cable Placement}}
& INT & 0/10 \\
& OPT & 9/10 \\
\midrule
\multirow{2}{*}{\textrm{Trefoil Knot}}
& INT & 0/10\\
& OPT & 9/10\\ 
\midrule
\multirow{2}{*}{\textrm{Reef Knot}}
& INT & 0/10\\
& OPT & 7/10\\ 
\bottomrule
\end{tabular}
\label{tab:knot}
\end{table}

Therein, we can see the potential of the deployment scheme in high-level robotic tasks like knot tying. In future work, the optimal deployment scheme will be incorporated with the perception system to automatically tie any prescribed knots with the robotics system.

\section{Conclusion}
\label{sec:conclusion}

In this article, we have introduced a novel deployment scheme that allows for robust and accurate control of the shape of DLOs using a single manipulator. Our framework integrates techniques from various disciplines, including physical simulation, machine learning, and scaling analysis, and has been demonstrated to be highly effective in real robotic experiments. Our results highlight the advantages of incorporating physics into robotic manipulation schemes and showcase impressive performance on complex tasks such as writing letters with elastic rods, cable placement, and tying knots.

Looking to the future, we plan to leverage the precision and efficiency of our deployment scheme to tackle some high-level robotic tasks systematically, for example, robotic knot tying. While exact shape control is not strictly required during such manipulations, our deployment scheme offers sufficient accuracy and efficiency to design the configurations of the middle states of a manipulated DLO, which is essential for robots to successfully tie complex knots. We also aim to explore the use of generalized problem formulations and data-driven control schemes, such as reinforcement learning, to develop more flexible and adaptive solutions to the challenges of robotic manipulation. By continuing to push the boundaries of robotic manipulation, we hope to advance the state-of-the-art in this field and enable new and exciting applications of robotic technology.

\end{document}